\documentclass[letterpaper]{article} 
\usepackage[preprint]{aaai2027}  
\usepackage[hyphens]{url}  
\usepackage{graphicx} 
\usepackage{natbib}  
\usepackage{caption} 
\usepackage{algorithm}
\usepackage{algorithmic}
\usepackage{amsmath}
\usepackage[most]{tcolorbox}
\usepackage{newfloat}
\usepackage{listings}
\DeclareCaptionStyle{ruled}{labelfont=normalfont,labelsep=colon,strut=off} 
\floatstyle{ruled}
\newfloat{listing}{tb}{lst}{}
\floatname{listing}{Listing}

\usepackage{booktabs}

\usepackage{url}
\usepackage{graphicx}
\usepackage{booktabs}
\usepackage{multirow}
\usepackage{color}
\usepackage{pifont}
\usepackage{subcaption} 
\usepackage{float}
\usepackage{colortbl}

\title{HomeSafeBench: Benchmarking Embodied Vision-Language Models in Free-Exploration Home Safety Inspection}
\author{
    Jiashu Yao\textsuperscript{\rm 1}\equalcontrib,
    Haoyu Wen\textsuperscript{\rm 1}\equalcontrib,
    Siyuan Gao\textsuperscript{\rm 1}\equalcontrib,
    Yuhang Guo\textsuperscript{\rm 1}\corresponding,
    Zeming Liu\textsuperscript{\rm 2},
    Heyan Huang\textsuperscript{\rm 1}
}
\affiliations{
    \textsuperscript{\rm 1}Beijing Institute of Technology\\
    \textsuperscript{\rm 2}Beihang University\\
    \{yaojiashu, wenhaoyu, gaosiyuan, guoyuhang, hhy63\}@bit.edu.cn\\
    zmliu@buaa.edu.cn 
}

\begin{document}

\maketitle

\begin{abstract}
Safety hazards in the home are a leading cause of preventable domestic injuries, motivating an automated inspector that actively explores a home and reports hazards before they cause harm. We introduce \textsc{HomeSafeBench}, the first benchmark for free-exploration home safety inspection with egocentric visual feedback, in which an embodied agent navigates a fully interactive 3D home, adjusts its viewpoint, and reports hazards purely from rendered first-person views. Built on the VirtualHome simulator, it covers five categories of common household hazards and comprises 1,000 human-validated inspection tasks. Evaluating a broad range of state-of-the-art Vision-Language Models (VLMs) reveals a large gap, where the best model reaches only about $34.7\%$ F1, far below the $98.0\%$ of a human inspector. Moreover, precision far exceeds recall across models, revealing a systematic tendency to under-report hazards that reflects a shared deficiency in risk recognition. To close this gap at low cost, we propose CueBack, an offline data-construction method that exploits the clue-precedes-confirmation structure of inspection, backtracking a privileged trajectory to the earliest frame where a hazard cue becomes visible and rewriting it into executable supervision. Fine-tuning a 4B-size VLM on CueBack-constructed data raises the average F1 from $18.7\%$ to $45.3\%$ on an out-of-distribution test set, surpassing the strongest closed-source model performance $34.7\%$. 
The benchmark, training dataset, and code are available at \url{https://github.com/BITHLP/HomeSafeBench}.
\end{abstract}

\section{Introduction}

Safety hazards in the home, such as flammable objects left beside an active stove or sharp tools within a child's reach, are a leading cause of preventable domestic injuries \citep{stewart2001home, josephson1991home, goldstick2022current}. Most of these incidents stem from ordinary human negligence and could be avoided if hazards were noticed and removed in time. However, continuous manual inspection of a home is tedious and impractical for residents to perform reliably. This motivates an automated inspector, an embodied agent that actively moves through a home, visually examines the environment, and reports safety hazards before they cause harm. Recent Vision-Language Models (VLMs) \citep{alayrac2022flamingo, liu2023visual, wang2024qwen2, Qwen2.5-VL} have enabled embodied agents to perform various practical tasks such as visual exploration, navigation, and embodied question-answering \citep{duan2022survey, chen2019learning, batra2020objectnav, ye2021auxiliary, zhao2025embodied}, especially within home environments \citep{yin2024safeagentbench, liu2024exploring}. Consequently, the automation of safety inspections using VLM agents is a promising new area of research.

\begin{figure*}[t]
    \centering
    \includegraphics[width=1.0\linewidth]{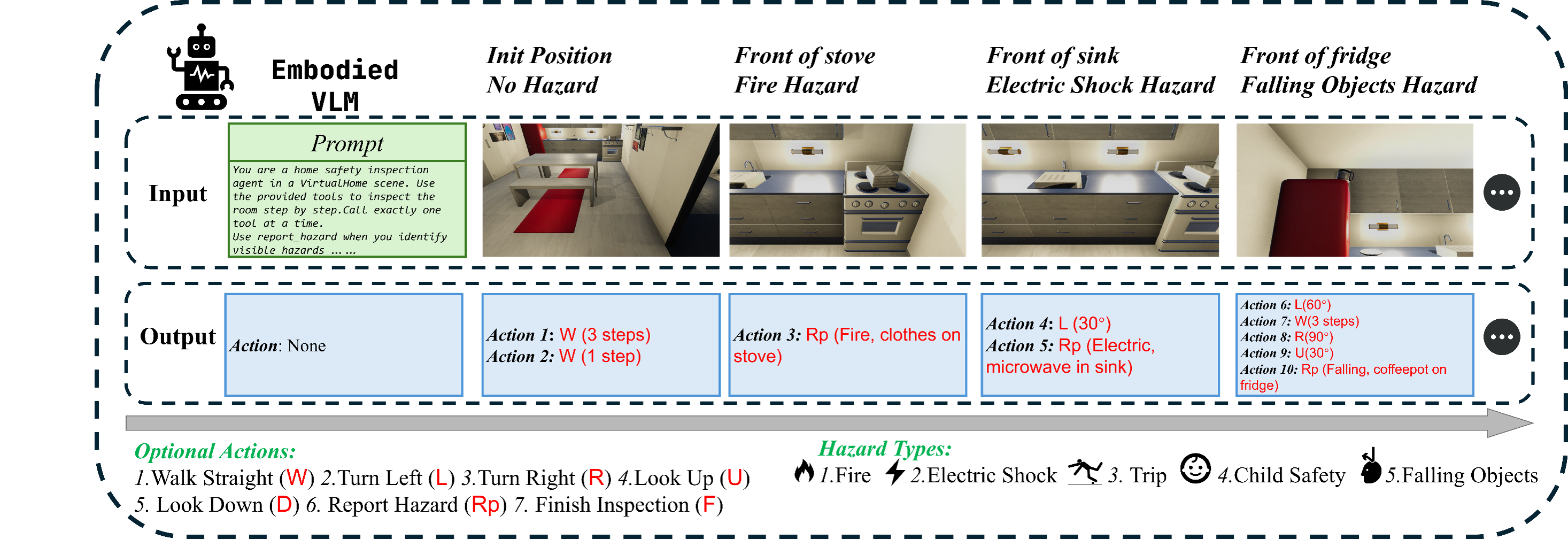}
    \caption{Schematic diagram of the free-exploration home safety inspection. VLM agents are tasked with identifying the objects that pose a safety hazard given the first-person perspective observation from the environment, and selecting the next action from the action list to iteratively inspect the entire room.}
    \label{fig:why-intro}
\end{figure*}

Realizing this vision requires a benchmark that reflects how home inspection actually happens. However, benchmarks dedicated to the household safety task either abstract the scene into text and discard the visual cues, or restrict the agent to a fixed viewpoint without free exploration. Home safety inspection, in contrast, is inherently active, requiring the agent to explore, look around, and approach suspicious regions from its own egocentric views under partial observability. To close this gap, we introduce \textsc{HomeSafeBench}, the first benchmark for free-exploration home safety inspection with egocentric visual feedback. Built on the VirtualHome simulator \citep{puig2018virtualhome}, \textsc{HomeSafeBench} places an embodied agent in a fully interactive 3D home and requires it to navigate, adjust its viewpoint, and report hazards purely from rendered first-person views, as shown in Figure \ref{fig:why-intro}. The benchmark covers five categories of common household hazards, i.e., fire, electric shock, falling object, trip, and child safety, and is constructed through a four-stage pipeline combining human annotation, rule-based single-hazard generation, VLM-based filtering, and data-level task composition. The resulting test set contains 1,000 inspection tasks with 3,000 hazard instances spanning 1 to 5 hazards per task, and human validation confirms that the tasks are navigable and their hazards reliably identifiable from agent views.

Evaluating a broad range of state-of-the-art VLMs on \textsc{HomeSafeBench} reveals a significant and consistent gap. While a human inspector reaches $98\%$ F1 on a subset, the best model reaches only about $35\%$, and open-source models trail even further. Across models we observe a systematic pattern in which precision far exceeds recall, showing that current VLMs overlook hazards, as they flag only the most obvious cases while overlooking the majority, especially falling-object and child-safety hazards. This under-reporting is a shared, systematic deficiency rather than a model-specific artifact. The bottleneck is thus a lack of specialized risk recognition, indicating that current VLMs lack the hazard awareness the safety inspection task demands and motivating targeted training to close the gap.

Improving this awareness through online reinforcement learning inflates the simulation cost, due to the heavy Unity-based simulator and the weak base agent performance. We instead offer a simulator-free training dataset to the community that enables a 4B-size open-source model to surpass SOTA closed-source ones. We propose CueBack, a lightweight data-construction method that turns a privileged trajectory into executable supervision. Our key observation is that inspection has a \emph{clue-precedes-confirmation} structure, where a hazard is first hinted at from a distance and only later confirmed up close. Given a privileged action trajectory, CueBack backtracks along the trajectory to the earliest egocentric frame in which the hazard cue becomes visible, and rewrites the reasoning accordingly, framing the earlier steps as exploration toward an unchecked high-risk region and the later steps as approaching to confirm a suspected hazard. This simple approach keeps every supervision signal executable without leaking privileged information. Fine-tuning a small Qwen3-VL-4B model on CueBack-constructed data lets it significantly outperform the strongest closed-source models on an out-of-distribution test set with disjoint scenes and hazard objects, raising the average F1 from $18.7\%$ to $45.3\%$, outperforming the strongest closed-source model ($34.7\%$) by a significant margin. Our contributions are listed below.
\begin{itemize}
    \item We introduce \textsc{HomeSafeBench}, the first benchmark for free-exploration home safety inspection in which an embodied agent actively navigates a 3D home and reports hazards from egocentric visual feedback, covering five hazard categories with 1{,}000 human-validated tasks.
    \item We conduct a comprehensive evaluation of state-of-the-art commercial and open-source VLMs and reveal that their failures stem from a systematic deficiency in hazard recognition, with especially low recall on falling-object and child-safety hazards.
    \item We propose CueBack, a lightweight offline data-construction method exploiting the clue-precedes-confirmation structure of inspection, which enables a 4B-size VLM to surpass leading closed-source models and other baselines on an out-of-distribution test set. 
    We release the benchmark, generated training dataset, and code at \url{https://github.com/BITHLP/HomeSafeBench}.
\end{itemize}

\section{Related Work}
\label{sec:related-work}
\subsection{Free Exploration for Embodied Agents}

Embodied agents often act under partial observability, where a single egocentric observation provides only incomplete evidence about the surrounding environment. 
This property has motivated work on embodied navigation, question answering, and free exploration \citep{anderson2018vision,das2018embodied,sripada2024scene,koo2025toward,yang2025embodiedbench}. 
Vision-and-language navigation evaluates whether an agent can follow natural-language instructions in visually grounded environments \citep{anderson2018vision}, while embodied question answering requires an agent to move through an environment and gather evidence for answering questions about unseen regions \citep{das2018embodied}. 
More directly related to our setting, active perception and free-exploration methods study how agents select informative viewpoints to improve scene understanding \citep{sripada2024scene,koo2025toward}.

These studies establish free exploration as a core capability for embodied agents, but their task objectives are typically navigation, question answering, object search, or general embodied reasoning. 
Home safety inspection differs in that the agent must decide where to inspect and report an initially unknown set of risk-bearing object-location configurations.

\subsection{Safety in Embodied AI}

Safety has become an important topic in embodied AI \citep{liu2024exploring,zhang2024badrobot,zhang2024safeembodai, huang2025framework, son2025subtle}. 
One line of work evaluates whether embodied agents themselves behave safely, such as refusing hazardous instructions \citep{ying2025agentsafe}, avoiding unsafe plans \citep{yin2024safeagentbench, zhu2024earbench}, or satisfying process-level physical constraints \citep{yang2026saferelbench}. These benchmarks mainly focus on risks induced by instructions, plans, or task execution processes. Another line of work uses embodied agents to address human safety-related problems \citep{li2025avd2, zhou2024hazard, hassan2024coherence}. The HAZARD Challenge studies decision making in dynamically changing disaster environments such as fire, flood, and wind \citep{zhou2024hazard}, while M-CoDAL focuses on multimodal safety dialogue and intervention for embodied agents \citep{hassan2024coherence}. 
These works shift the focus from the safety of the agent's own behavior to safety risks that affect humans in the surrounding environment.

Our home safety inspection task falls into this second line of work, as it focuses on risks that affect humans in everyday home environments.

\subsection{Home Safety and Hazard Detection}
Home safety has recently been studied in embodied and vision-based benchmarks, but existing settings do not fully capture free-exploration home safety inspection.
SafetyDetect is the closest prior work to our hazard-discovery objective, as it detects unsafe or anomalous household states such as unsafe object placements and unsanitary conditions \citep{mullen2024don}.
However, it mainly relies on scene graphs and object-relation representations, where the visual evidence needed for hazard recognition is omitted.

Other benchmarks retain visual observations or embodied interaction, but evaluate agents under prescribed tasks rather than free exploration.
HomeGuard provides task instructions and visual context for judging contextual risks in household tasks \citep{lu2026homeguard}. 
SafetyALFRED augments ALFRED-style kitchen tasks with hazards and evaluates step-conditioned safety decisions, including hazard-mitigation actions \citep{torres2026safetyalfred}.
IS-Bench evaluates whether agents can avoid or mitigate risks while executing daily household tasks through interaction with the environment \citep{lu2026bench}. 
Video-based safety benchmarks evaluate unsafe action or runtime risk detection from household or egocentric videos \citep{pu2026homesafe,panpatil2026egosafetybench}. 
Together, these settings evaluate safety under prescribed tasks, pre-recorded trajectories or step-conditioned decisions, rather than free exploration for discovering an unknown set of hazards.

In contrast, \textsc{HomeSafeBench} evaluates free-exploration home safety inspection. 
The agent must explore a partially observable home, decide its own observation, and report an unknown set of hazards.

\section{Benchmark}

\subsection{Task Definition}
We propose a home safety hazard inspection task in which an embodied agent actively navigates a simulated 3D home environment to identify and report safety hazards. 
Following real-world home safety guidelines, we define five categories of common household hazards in our benchmark. 
Each category represents a specific configuration of item placement that poses a safety risk.

\begin{itemize}
    \item \textbf{Fire hazards}: Flammable materials are located close to active or potential heat sources. 
    Examples include curtains or stacks of paper placed next to a lit stove, and a pile of dry cloth near a burning candle.

    \item \textbf{Electric shock hazards}: Appliances or power devices in contact with water, which may cause electric shock or short circuits.
    Examples include an appliance in a sink.

    \item \textbf{Falling object hazards}: Items positioned in a way that they may fall from height and cause injury or damage. 
    Examples include a coffee pot placed at the edge of a refrigerator, or a box positioned at the edge of a shelf.

    \item \textbf{Trip hazards}: Objects or clutter on the floor that could cause someone to stumble or lose balance during normal movement. 
    Examples include a banana left in a hallway.

    \item \textbf{Child safety hazards}: Placement of dangerous or harmful items within easy reach of a child. 
    Examples include sharp kitchen knives placed on a TV stand.
\end{itemize}
Formally, let the initial state be denoted by $s_0$, with a ground-truth hazard set $\mathcal{H}$. 
At each discrete time step $t$, the agent receives an observation $o_t$ and selects an action $a_t \in \mathcal{A}$ based on its history $h_t$:
\begin{align}
    a_t \sim \pi(\cdot | h_t).
\end{align}
The action space $\mathcal{A}$ includes navigation-related inspection actions, such as movement, turning, and viewpoint adjustment, as well as hazard-reporting and termination actions.
Navigation-related actions update the agent pose or viewpoint, whereas hazard-reporting actions update the reported hazard set without changing the environment state.
All actions are implemented via tool calls.
After executing a sequence of actions $\{a_0, a_1, \dots, a_{T-1}\}$ within a step budget $T$, the final reported hazard set is denoted as $\hat{\mathcal{H}}$. 
Task performance is evaluated by comparing the reported hazards $\hat{\mathcal{H}}$ against the ground-truth hazards $\mathcal{H}$ using precision, recall, and F1 under two matching protocols.
Category-F1 measures whether the agent reports the correct hazard categories, while Hazard-F1 uses stricter instance-level matching that requires both the hazard category and the involved objects to match a ground-truth hazard.
To account for semantic variation in VirtualHome object names, hazard-level matching is performed by a VLM judge. 
More details are shown in the Appendix.

\begin{figure*}[t]
    \centering
    \includegraphics[width=\textwidth]{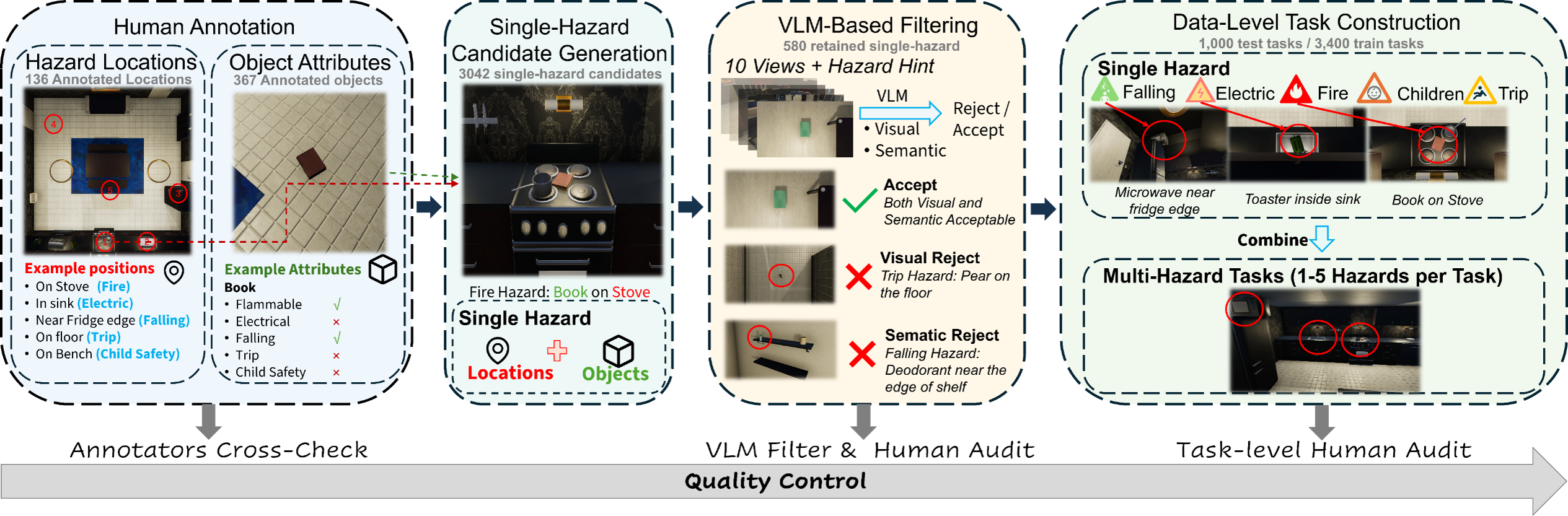}
    \caption{Four-stage pipeline for constructing the HomeSafeBench benchmark.}
    \label{fig:dataset-pipeline}
\end{figure*}

\subsection{Dataset Construction}
\label{sec:construction}
\textsc{HomeSafeBench} is built on VirtualHome through a four-stage pipeline that combines human annotation, single-hazard candidate generation, VLM-based filtering, and data-level task composition, as illustrated in Figure~\ref{fig:dataset-pipeline}.

\paragraph{Human annotation.} 
We first annotate potential hazard locations and object safety attributes in 12 room scenes from three VirtualHome environments. 
These annotations jointly specify where hazards may occur and which objects can instantiate them.
For hazard locations, annotators identify spatial regions that may become unsafe after an appropriate object is placed there, such as the top of a refrigerator, the interior of a sink, or the area near a stove.
Each location is associated with exactly one hazard type: fire, electric shock, falling object, trip, or child safety.
Two annotators each perform the initial annotation for six rooms across the three environments and then cross-check the other annotator's results case by case, removing low-risk locations and retaining only consensus annotations.
This process yields 136 annotated hazard locations.
For object attributes, two annotators independently determine whether common objects are flammable, electrical, likely to induce tripping or falling-object hazards, or unsafe for children, with multiple attributes allowed for each object.
Disagreements are adjudicated by a third annotator, resulting in 367 objects with safety-related attributes.

\paragraph{Single-hazard candidate generation.}
We generate single-hazard candidates by matching the hazard type of an annotated location with compatible object attributes and placing a matched object at the corresponding location.
For example, paper can be placed near a stove to create a fire hazard, whereas a glass can be placed on top of a refrigerator to create a falling-object hazard.
The rule-based pairing and placement procedure produces 3,042 single-hazard candidates.

\paragraph{VLM-based filtering.}
Rule compatibility alone does not guarantee that a generated hazard is visually observable or semantically convincing, so we use a VLM to filter every candidate.
For each candidate, the VLM receives the intended hazard type and involved objects as a textual hint, together with ten rendered views consisting of one frontal view, one top-down view, and eight diagonal views captured at upper and lower elevations.
The VLM jointly assesses visual clarity, requiring the relevant objects and spatial relations to be observable, and semantic validity, requiring the object-location configuration to constitute the intended household hazard.
Only candidates satisfying both criteria are retained.
This process retains 580 of the 3,042 single-hazard candidates, corresponding to a retention rate of 19.07\%.
\paragraph{Data-level task construction.}
Finally, we combine compatible single-hazard cases according to the composition rules to construct inspection tasks with different numbers of hazards.
Each task contains $N=1$--$5$ hazards, and only single-hazard cases that can validly coexist in the same environment are composed.
The resulting \textsc{HomeSafeBench} test set contains 1,000 tasks spanning different hazard counts and room configurations.

\paragraph{Dataset statistics.}
The final test split is built from four held-out room scenes from one VirtualHome environment.
It contains 1,000 inspection tasks with 3,000 hazard instances in total. 
The tasks are uniformly distributed by hazard count, with 200 tasks for each $N \in \{1,2,3,4,5\}$ and 250 tasks per room.
Table~\ref{table:dataset_comparison} compares \textsc{HomeSafeBench} with existing safety-related benchmarks. More details are shown in the Appendix.

\begin{table}[t]
\centering
\setlength{\tabcolsep}{2mm}
\begin{tabular}{@{}lcccccc@{}}
\toprule
\textbf{Dataset} & \textbf{S} & \textbf{H} & \textbf{VI} & \textbf{FE} & \textbf{MDS} & \textbf{MT} \\
\midrule
SafetyDetect
      & 1,000 & 3  &  {\ding{56}}   &  {\ding{56}} &  {\ding{56}}   &  {\ding{56}} \\
M-CoDAL
      & 908 & 16   &  {\ding{52}} &  {\ding{56}} &  {\ding{56}}   &  {\ding{56}} \\
SafeAgentBench
      & 750 & 10   &  {\ding{52}} &  {\ding{56}} &  {\ding{56}}   &  {\ding{56}} \\
SafePlan-Bench
      & 2,027 & 8  &  {\ding{56}}   &  {\ding{56}} &  {\ding{56}}   &  {\ding{56}} \\
EmbodyGuard
      & 942 & 12   &  {\ding{56}}   &  {\ding{56}} &  {\ding{56}}   &  {\ding{52}} \\
IS-Bench
      & 388 & 10   &  {\ding{52}} &  {\ding{56}} &  {\ding{56}}   &  {\ding{52}} \\
\midrule
\textbf{\textsc{HomeSafeBench}}
      & 1,000 & 5 &  {\ding{52}} &  {\ding{52}} &  {\ding{52}} &  {\ding{52}} \\
\bottomrule
\end{tabular}
\caption{Comparison of \textsc{HomeSafeBench} with existing safety-related datasets \citep{mullen2024don,hassan2024coherence,yin2024safeagentbench,huang2025framework,son2025subtle,lu2026bench}. S/H denotes samples / hazard categories. VI, FE, MDS, and MT denote visual interaction, free exploration, multiple dangerous scenarios, and multi-turn interaction.}
\label{table:dataset_comparison}
\end{table}

\subsection{Human Quality Check}
\label{sec:quality-control}
We conduct human validation at two levels to assess the quality of the generated benchmark data. 
First, to verify the reliability of VLM-based filtering, a human reviewer manually audited 116 of the 580 VLM-retained single-hazard candidates (20.0\%), assessing each candidate in terms of semantic validity and visual clarity, and no errors were found in this sampled audit. Second, to validate the final data-level inspection tasks, we sampled 50 of the 1,000 composed tasks (5.0\%), with 10 tasks for each $N \in \{1,2,3,4,5\}$. An independent human evaluator, who was not involved in any stage of data annotation or task construction, performed each task under the standard benchmark protocol, without access to the ground-truth hazard annotations or the number of hazards. Across the sampled tasks, the evaluator achieved 98.00\% micro-F1, indicating that the composed tasks are navigable and that their hazards can be reliably identified from agent-view observations.

\section{Method}

To improve hazard awareness without the prohibitive cost of coupling the heavy Unity-based simulator into an online training loop, we construct an offline supervised fine-tuning (SFT) dataset. By releasing this simulator-free dataset, we also aim to make competitive home safety inspection attainable for practitioners without access to large-scale simulation or proprietary models, enabling a locally deployable agent that approaches commercial-grade performance.

The central challenge is that teacher trajectories are contaminated by \emph{privileged information}, as the teacher acts as if it already knows where each hazard is, whereas the deployed agent must rely on egocentric observations alone. Naively imitating them thus harms generalization to unseen scenes and hazards. Our pipeline addresses this in three stages. We first generate training environments while ensuring that their scenes and hazard objects are disjoint from the test set, then label the executable inspection actions within each environment, and finally generate observation-grounded reasoning that removes privileged-information leakage.

\subsection{Task Construction}
We follow the same task construction procedure for both the training and test sets. 
The training set is constructed from two VirtualHome environments, comprising eight room scenes across four room types. 
The test set is constructed from a held-out environment containing four room scenes. 
The two splits are disjoint in environments, scene layouts, and object instances, thereby preventing data leakage and enabling evaluation in an out-of-domain environment. 
The training split contains 3,400 tasks, including 371 single-hazard tasks and 658, 657, 657, and 1,057 tasks with two, three, four, and five hazards, respectively.

\subsection{Action Trajectory Annotation and Synthesis}

To obtain executable training trajectories, we combine manual annotation with automatic trajectory synthesis. 
Since different tasks instantiated in the same room share the same spatial structure and set of potential hazard locations, we construct a reusable room-level directed observation graph instead of manually annotating a complete action sequence for every sample. 
Each node represents a validated observation viewpoint and records the hazard locations that can be inspected from it, while each directed edge represents an executable navigation sequence between two viewpoints.

Given a training sample, we select observation viewpoints that cover its target hazards and connect them through the annotated graph. 
The resulting navigation actions are combined with the corresponding inspection and hazard-reporting actions to form candidate trajectories, which are replayed in the simulator. 
After replay-based validation, we obtain 3,158 executable action trajectories. 
Further details on action trajectory annotation and synthesis are provided in the Appendix.

\subsection{Clue-Precedes-Confirmation Reasoning}

Given an action trajectory $\tau = \{(o_t, a_t)\}_{t=0}^{T-1}$ obtained from the previous stage, our goal is to attach to each step a reasoning trace $r_t$ that explains \emph{why} $a_t$ is taken. Our construction exploits a structural property specific to home safety inspection, which we term \emph{clue precedes confirmation}, which means a hazard is rarely recognizable at the start of a trajectory, but leaves a visual clue that becomes observable from a distance well before the agent is close enough to confirm it. Consequently, each hazard-directed segment of a trajectory admits a natural split point, i.e., the earliest step at which the corresponding clue enters the agent's egocentric view. For each hazard $h$ associated with a segment, we scan its observations $\{o_t\}$ in temporal order and use a vision-language detector to locate the first frame $k_h$ in which the clue of $h$ is visible.

The split point $k_h$ partitions the segment into two phases, each supervised with a distinct reasoning style. Before $k_h$, when no clue is yet visible, the reasoning is written in an exploration style. It justifies navigation actions as coverage-driven decisions to inspect unchecked, high-prior regions of the room, without ever naming the hazard that lies ahead. From $k_h$ onward, when the clue has become observable, the reasoning switches to a confirmation style. It acknowledges the visible clue, hypothesizes the potential risk it suggests, and justifies approaching to obtain a closer view before committing to a report. The final reporting action is then justified by the now clearly observed hazard. This phase-dependent rewriting ensures that every reasoning trace is grounded in the current observation, and exploration steps never leak the identity or location of a not-yet-visible hazard, while confirmation steps reference only clues that are genuinely present in the agent's view.
Further details on CueBack reasoning generation and implementation are provided in the Appendix.

\begin{table*}[t]
\centering
\setlength{\tabcolsep}{1.15mm}
\begin{tabular}{lcccccccccccccccccc}
\toprule
\multirow{2}{*}{\textbf{Model}} & \multicolumn{3}{c}{\textbf{Average}} & \multicolumn{3}{c}{\textbf{Trip}} & \multicolumn{3}{c}{\textbf{Falling}} & \multicolumn{3}{c}{\textbf{Fire}} & \multicolumn{3}{c}{\textbf{Children}} & \multicolumn{3}{c}{\textbf{Electric}} \\
\cmidrule(lr){2-4} \cmidrule(lr){5-7} \cmidrule(lr){8-10} \cmidrule(lr){11-13} \cmidrule(lr){14-16} \cmidrule(lr){17-19}
 & P & R & F1 & P & R & F1 & P & R & F1 & P & R & F1 & P & R & F1 & P & R & F1 \\
\midrule
Human & 98.0& 98.0& 98.0& 96.8& 100 & 98.4 & 97.1 & 94.3 & 95.7 &  100 & 100 & 100 & 100 & 100 & 100 & 100 & 87.5 & 93.3 \\

\midrule
\rowcolor{gray!20} \multicolumn{19}{c}{\it Commercial VLMs}\\
Seed-2.0-Pro & 41.4 & 29.9 & 34.7 & \underline{72.6} & 39.8 & 51.4 & 15.9 & 11.3 & 13.2 & 56.3 & \underline{43.3} & \underline{49.0} & 15.7 & 17.7 & 16.6 & 17.5 & 19.0 & 18.2 \\
GPT-5.5 & 30.4 & \underline{31.2} & 30.8 & 46.4 & \textbf{57.1} & 51.2 & 13.3 & 14.5 & 13.8 & 40.3 & 9.0 & 14.7 & 9.4 & 14.8 & 11.5 & 26.6 & \textbf{28.7} & \underline{27.6} \\
GLM-5V-Turbo & \textbf{58.9} & 17.8 & 27.3 & \textbf{83.2} & 37.0 & 51.2 & 21.5 & 4.1 & 6.9 & 54.0 & 8.1 & 14.2 & 2.8 & 0.3 & 0.6 & 13.6 & 8.6 & 10.6 \\
Qwen-3.6-Plus & 45.7 & 19.3 & 27.1 & 71.0 & 35.5 & 47.4 & 10.0 & 2.5 & 4.0 & 55.6 & 18.2 & 27.4 & 8.7 & 3.9 & 5.4 & 12.5 & 10.9 & 11.7 \\
Kimi-K2.5 & 35.8 & 14.0 & 20.1 & 53.6 & 20.8 & 29.9 & 15.6 & 2.8 & 4.7 & \underline{56.9} & 17.2 & 26.4 & 9.0 & 5.6 & 6.9 & 16.7 & 20.7 & 18.5 \\

\midrule
\rowcolor{gray!20} \multicolumn{19}{c}{\it Open-Source VLMs}\\
Qwen3-VL-8B & 40.2 & 14.2 & 21.0 & 41.5 & 34.8 & 37.8 & 37.0 & 1.3 & 2.6 & 0.0 & 0.0 & 0.0 & 11.1 & 0.3 & 0.6 & 0.0 & 0.0 & 0.0 \\
Qwen3-VL-4B & 47.2 & 11.6 & 18.7 & 68.6 & 26.2 & 37.9 & 3.7 & 0.7 & 1.1 & 13.5 & 1.7 & 3.1 & 26.7 & 3.9 & 6.9 & 37.9 & 6.5 & 11.1 \\
Qwen3-VL-A3B & 22.7 & 1.4 & 2.6 & 44.9 & 2.9 & 5.5 & 0.0 & 0.0 & 0.0 & 26.7 & 0.7 & 1.4 & 8.3 & 0.3 & 0.6 & 3.3 & 1.1 & 1.7 \\

\midrule
\rowcolor{gray!20} \multicolumn{19}{c}{\it Qwen3-VL-4B Trained with Our Data}\\
SFT-Action & 49.4 & 30.6 & 37.8 & 57.1 & \underline{52.4} & \underline{54.7} & 18.5 & 7.8 & 11.0 & 50.9 & 30.0 & 37.7 & \textbf{73.0} & 13.7 & 23.1 & 48.6 & 15.7 & 23.7 \\
SFT-Both & 55.7 & 30.4 & \underline{39.3} & 58.5 & 40.5 & 47.8 & \textbf{49.5} & \textbf{18.6} & \textbf{27.0} & 55.4 & 31.9 & 40.5 & \underline{50.3} & \textbf{26.9} & \textbf{35.0} & \textbf{73.5} & 14.4 & 24.0 \\
DPO & 40.6 & 22.7 & 29.1 & 50.7 & 46.4 & 48.4 & 6.7 & 0.6 & 1.0 & 9.1 & 1.6 & 2.7 & 17.6 & 20.7 & 19.0 & 0.0 & 0.0 & 0.0 \\
CueBack (ours) & \underline{58.3} & \textbf{37.1} & \textbf{45.3} & 63.5 & 49.3 & \textbf{55.5} & \underline{44.4} & \underline{14.9} & \underline{22.3} & \textbf{57.7} & \textbf{51.0} & \textbf{54.1} & 47.1 & \underline{23.6} & \underline{31.4} & \underline{69.6} & \underline{27.7} & \textbf{39.7} \\
\bottomrule
\end{tabular}%
\caption{Main results grouped by hazard type. The best and second-best performance are bold and underlined, respectively.}
\label{tab:homesafebench-main}
\end{table*}

\section{Experiments}

\subsection{Data Generation Settings}
Both the benchmark and the training data rely on a VLM for automated generation, for which we use GPT-5.5 throughout. For the benchmark, the VLM performs the filtering stage that decides whether a single-hazard candidate is visually clear and semantically valid. For the training data, the same model is additionally used to locate the split point at which a hazard clue first becomes visible along a trajectory, and to generate the observation-grounded reasoning for each step. All prompts used for filtering, split-point localization, and reasoning generation are provided in the Appendix.

\subsection{Experimental Settings}

\paragraph{Agent design.}
We wrap the low-level operations of the VirtualHome engine as a set of native tools that the agent invokes to move, turn, adjust its viewpoint, and report hazards, as illustrated in Figure~\ref{fig:why-intro}. At each step, the agent receives its first-person view rendered at a resolution of $640 \times 360$, and the full interaction history is retained across steps. Each episode is capped at a maximum of 20 steps.

\paragraph{Evaluated models.}
We evaluate a broad range of state-of-the-art VLMs. For commercial VLMs, we test Doubao-Seed-2.0-Pro, GPT-5.5, GLM-5V-Turbo, Qwen-3.6-Plus, and Kimi-K2.5. For open-source models, we evaluate the Thinking variants of Qwen3-VL-8B, Qwen3-VL-4B, and Qwen3-VL-30B-A3B. For brevity, the \mbox{-Thinking} suffix is omitted in the tables. All models are run with reasoning enabled, using a temperature of $0.6$ and top-$p$ of $0.95$.

\paragraph{Training implementation.}
All training is performed on Qwen3-VL-4B-Thinking. We do not exhaustively tune hyperparameters and adopt a single configuration across all methods. We use LoRA with rank $16$, $\alpha=32$, and dropout $0.05$, applied to all attention and feed-forward layers. Models are trained with a maximum sequence length of 16K tokens, a batch size of $16$, and the AdamW optimizer with a learning rate of $1\times10^{-4}$ under a cosine schedule with $100$ warmup steps, for a single epoch.

\paragraph{Baselines.}
We compare CueBack against three baselines that share the same action trajectories but differ in how supervision is constructed. \emph{SFT-Action} uses only the executable actions without any reasoning. \emph{SFT-Both} augments the actions with reasoning, but instead of our clue-precedes-confirmation design, it directly prompts GPT-5.5 to generate a rationale given the trajectory prefix and the current action. \emph{DPO} is an annotation-free offline preference method. For each task we sample four rollouts from Qwen3-VL-4B-Thinking and construct a preference pair by preferring the best rollout over the worst, ranking primarily by F1 and breaking ties toward fewer steps.

\paragraph{Metrics.}
The evaluation metrics are defined in the Benchmark section. In the main text, we report hazard-level precision, recall, and F1 based on a GPT-5.5 judge, as this hazard-level matching is more accurate and less prone to overestimation. The category-level metrics, which do not rely on an LLM judge, are reported in the Appendix.

\subsection{Main Results}

\paragraph{Existing VLMs struggle with home safety inspection.}
As shown in Table~\ref{tab:homesafebench-main}, all evaluated models fall far short of the human inspector, whose F1 reaches $98.0$, while the best closed-source model, Doubao-Seed-2.0-Pro, attains only $34.7$ and open-source models trail further behind. Beyond the low overall scores, a consistent pattern emerges across models: precision substantially exceeds recall (e.g., $58.9$ vs. $17.8$ for GLM-5V-Turbo and $45.7$ vs. $19.3$ for Qwen-3.6-Plus), revealing a systematic tendency to overlook hazards. The models flag only the most salient cases, typically trip hazards, while overlooking the majority, with especially weak performance on falling-object and child-safety hazards. This exposes a shared deficiency in risk recognition.

\paragraph{Our data substantially improves inspection ability at low cost.}
Since the training environments and hazard objects are disjoint from the test set, the gains reported here reflect genuine generalization rather than information leakage. Fine-tuning Qwen3-VL-4B-Thinking on our data yields large improvements over the corresponding base model, whose F1 is only $18.7$. Notably, even the simplest usage of our data, SFT-Action, which supervises executable actions without any reasoning, lifts the 4B model to $37.8$ F1, already surpassing the strongest closed-source model ($34.7$). This demonstrates that our simulator-free data alone is sufficient to bring a small open-source model to the level of top commercial systems.

\paragraph{CueBack is simple yet highly effective.}
Building reasoning on top of the actions further improves performance, and our clue-precedes-confirmation design is markedly more effective than a naive rationale. Compared with SFT-Both, the most directly comparable baseline that also pairs actions with reasoning but generates it without our design, CueBack improves F1 from $39.3$ to $45.3$ ($+6.0$), and it clearly outperforms the annotation-free DPO ($29.1$). Against the best closed-source model, CueBack raises F1 from $34.7$ to $45.3$ ($+10.6$) and lifts recall from $29.9$ to $37.1$, with substantial gains precisely on the hardest categories, e.g., child-safety F1 from $16.6$ to $31.4$ and electric-shock F1 from $27.6$ to $39.7$. These results show that a lightweight, simulator-free recipe can turn a small open-source model into a hazard inspector that significantly surpasses leading commercial VLMs.

\subsection{Analysis of Agent Behavior}

\begin{figure}[htbp]
    \centering
    \includegraphics[width=1.0\linewidth]{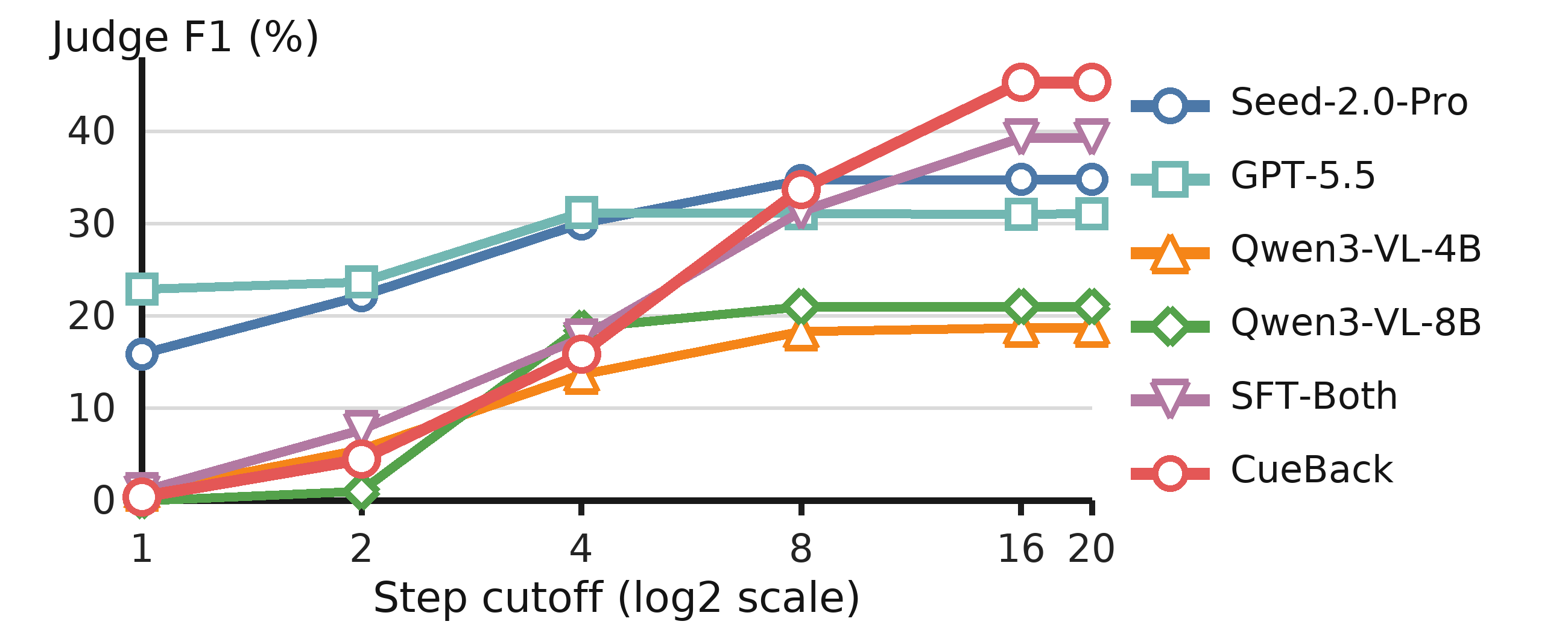}
    \caption{Judge F1 as a function of the step cutoff.}
    \label{fig:cutoff-f1}
\end{figure}

\begin{figure}[htbp]
    \centering
    \includegraphics[width=1.0\linewidth]{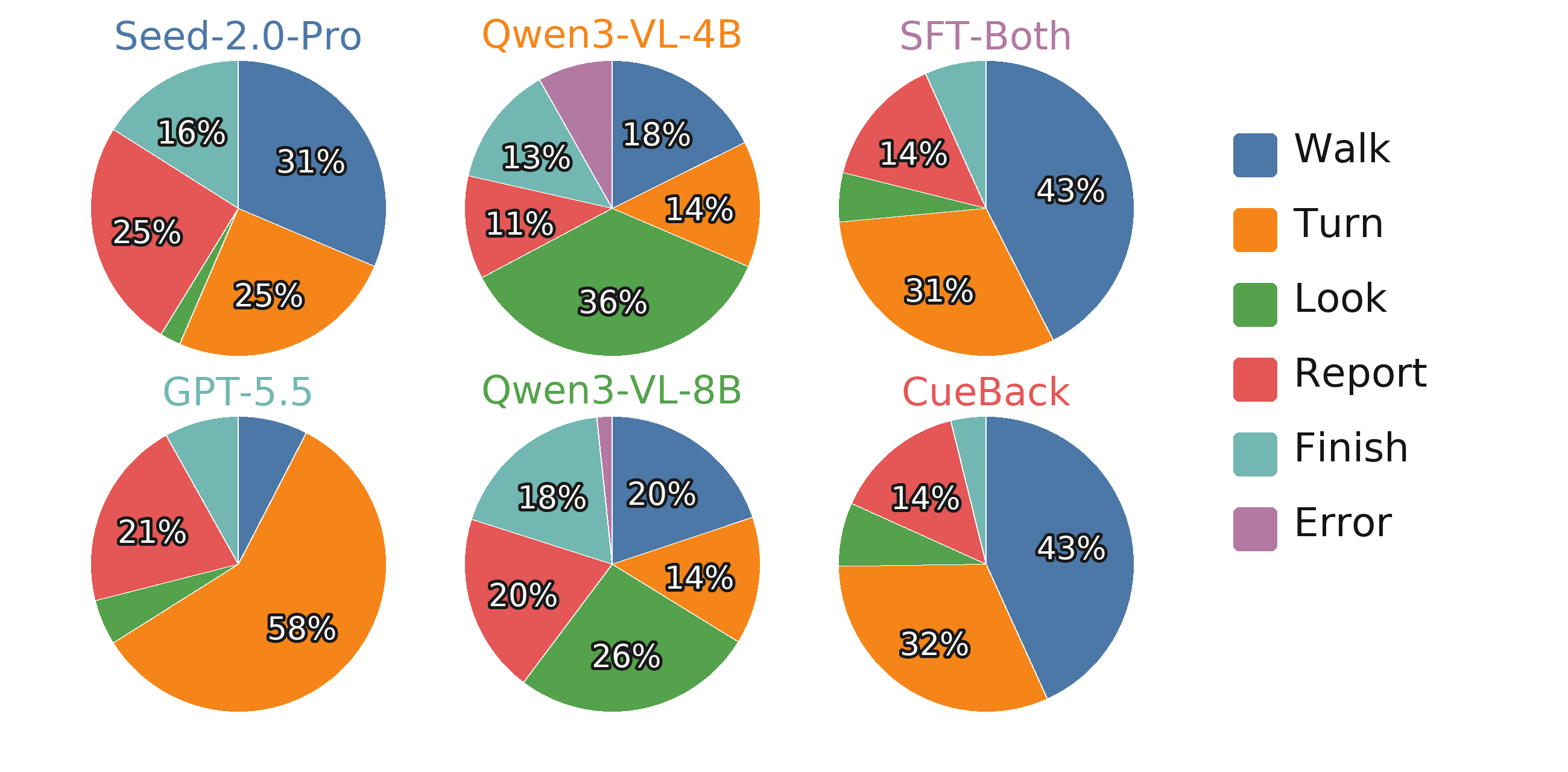}
    \caption{Distribution of executed action types for six representative agents over all inspection tests.}
    \label{fig:action-ratio}
\end{figure}

\paragraph{Trained models sustain inspection over longer horizons.}
Figure~\ref{fig:cutoff-f1} plots how F1 accumulates as more steps are allowed. Untrained models lack the ability to inspect persistently, as their curves rise sharply in the early steps and then quickly saturate. This is most pronounced for the strong commercial VLMs, which rapidly discover and report the salient hazards near the starting location but make little further progress afterwards, flattening well before the step budget is exhausted. In contrast, the model trained on our data keeps discovering and reporting hazards throughout the episode, with its curve continuing to climb in the later steps and eventually surpassing all others. This indicates that the recognition knowledge instilled by our data combines with continued exploration to turn additional viewpoints into additional correct reports, rather than stalling once the obvious cases are exhausted.

\paragraph{The bottleneck is recognition, not under-reporting.}
To explain this saturation, Figure~\ref{fig:action-ratio} examines the action composition of each agent. For the untrained models, report actions account for a substantial share of their behavior (21\%--25\% for GPT-5.5 and Seed-2.0-Pro), comparable to or higher than the trained models. Their limitation instead lies in exploration, as GPT-5.5 spends 58\% of its actions turning in place with very little walking, and the base open-source models devote large fractions to looking up and down, so all of them cover little ground and rarely bring new regions into view. The trained models behave differently, allocating far more actions to walking (43\% for CueBack) and thereby actively expanding the observed area. Combined with their stronger recognition of specific hazard types, this explains why untrained models overlook so many hazards and F1 plateaus early, whereas CueBack keeps improving.

\paragraph{Free exploration matters, but only if the agent can exploit it.}
To isolate the role of free exploration, we ablate the available actions in Table~\ref{tab:action-ablation-micro}: \emph{w/o walk} disables locomotion so the agent can only adjust its viewpoint from a fixed position, and \emph{report-only} further removes viewpoint control, reducing inspection to a static single view. On the one hand, free exploration is clearly necessary, as performance drops under both restrictions for all models. On the other hand, the magnitude of the drop reveals how differently each agent uses this freedom. The untrained closed-source models are far less affected, e.g., GPT-5.5 loses only $2.4$ and $7.8$, indicating that they gain little from free exploration and remain unable to fully exploit it.

\begin{table}[htbp]
\centering
\begin{tabular}{lccc}
\toprule
\textbf{Setting} & \textbf{CueBack} & \textbf{GPT-5.5} & \textbf{Seed-2.0-Pro} \\
\midrule
Original & 45.3 & 30.8 & 34.7 \\
w/o walk & 29.2 (-16.1) & 28.4 (-2.4) & 22.8 (-11.9) \\
report-only & 24.8 (-20.5) & 23.0 (-7.8) & 17.8 (-16.9) \\
\bottomrule
\end{tabular}
\caption{Ablation on available actions. Values are F1 scores.}
\label{tab:action-ablation-micro}
\end{table}

\section{Conclusion}

We introduced \textsc{HomeSafeBench}, the first benchmark for free-exploration home safety inspection with egocentric visual feedback, in which an embodied agent navigates a 3D home and reports hazards from first-person views. Evaluating a broad range of state-of-the-art VLMs reveals that they remain far behind humans, primarily due to a deficiency in hazard recognition. To address this, we proposed CueBack, a simulator-free offline data-construction method that exploits the clue-precedes-confirmation structure of inspection to turn privileged trajectories into executable supervision, enabling a 4B-size model to surpass the strongest closed-source models on out-of-distribution tasks. 
We release the benchmark, the training dataset, and all code to support future research on embodied home safety inspection.

\bibliography{references,original}

\clearpage

\appendix
\section{Benchmark Details}
\label{app:benchmark-details}
\subsection{Virtual Environment and Agent Interaction}
\label{app:tool-execution}
We implement HomeSafeBench in VirtualHome. 
At the beginning of each episode, the task scene is instantiated in VirtualHome, the agent is placed at a predefined initial pose, and an initial egocentric observation is provided.
The agent then interacts with the environment by invoking exactly one tool at each step. 
After the tool is executed, the environment returns the corresponding feedback and an updated observation for the next step.
The agent interacts with the environment through six semantic tools, whose interfaces and arguments are summarized in Table~\ref{tab:tool-schema}.

\begin{table}[h]
\centering
\small
\begin{tabular}{@{}ll@{}}
\toprule
Tool & Arguments \\
\midrule
\texttt{walk} &
$\texttt{steps}\in\{1,\ldots,5\}$ \\
\texttt{turn} &
\begin{tabular}[t]{@{}l@{}}
$\texttt{direction}\in\{\texttt{left},\texttt{right}\}$ \\
$\texttt{angle}\in\{30^\circ,60^\circ,\ldots,180^\circ\}$
\end{tabular} \\
\texttt{look\_up} &
$\texttt{angle}\in[5^\circ,75^\circ]$ \\
\texttt{look\_down} &
$\texttt{angle}\in[5^\circ,75^\circ]$ \\
\texttt{report\_hazard} &
One or more category--object reports \\
\texttt{finish\_inspection} &
A textual termination reason \\
\bottomrule
\end{tabular}
\caption{Semantic tool interface exposed to agents in HomeSafeBench.}
\label{tab:tool-schema}
\end{table}

The \texttt{walk} tool moves the agent forward by the specified number of steps. The \texttt{turn} tool rotates the agent in the specified direction by the given angle. Since VirtualHome provides native turning actions in increments of $30^\circ$, the supported rotation angles are restricted to multiples of $30^\circ$.

The \texttt{look\_up} and \texttt{look\_down} tools adjust only the egocentric camera without changing the agent position. After the corresponding observation is obtained, the camera returns to its default orientation. The \texttt{report\_hazard} tool appends the reported hazards to the cumulative prediction set without modifying the simulated scene, while \texttt{finish\_inspection} terminates the current episode. An episode also terminates when the predefined interaction budget is reached. The hazards accumulated before termination constitute the final prediction set $\hat{\mathcal{H}}$, which is used for evaluation.

\subsection{Evaluation Protocol}
\label{app:evaluation-protocol}

We compare the final prediction set $\hat{\mathcal{H}}$ with the ground-truth hazard set $\mathcal{H}$ using precision, recall, and F1. 
Since hazard reports can be matched to ground truth under different levels of strictness, we instantiate the metric under two matching protocols: category-level matching and hazard-level matching. 
Both protocols first determine the numbers of true positives, false positives, and false negatives, and then compute precision, recall, and F1 as
\begin{align}
    P = \frac{\mathrm{TP}}{\mathrm{TP}+\mathrm{FP}}, \quad
    R = \frac{\mathrm{TP}}{\mathrm{TP}+\mathrm{FN}}, \quad
    F1 = \frac{2PR}{P+R}.
\end{align}

\paragraph{Category-level matching.}
Category-level matching evaluates whether the agent reports the correct distribution of hazard categories, ignoring the associated object instances. 
Let $g_c$ and $p_c$ denote the number of ground-truth and predicted hazards of category $c$, respectively. We compute
\begin{align}
    \mathrm{TP}_{cat} &= \sum_c \min(g_c, p_c), \\
    \mathrm{FP}_{cat} &= \sum_c \max(0, p_c - g_c), \\
    \mathrm{FN}_{cat} &= \sum_c \max(0, g_c - p_c).
\end{align}
This metric captures whether the agent identifies the correct types of hazards, but it does not verify whether the reported object instance is correct.

\paragraph{Hazard-level matching.}
Hazard-level matching evaluates hazard reports at the instance level. 
We use a VLM judge rather than exact object-name matching because object names in VirtualHome may be coarse, ambiguous, or different from the way agents refer to visually observed objects.
The judge receives the predicted hazards and the ground-truth hazards.
Each predicted hazard contains a reported hazard category and associated objects, while each ground-truth hazard contains its category, associated objects, and semantic location. 
The judge performs one-to-one matching between predicted and ground-truth hazards: each predicted hazard can match at most one ground-truth hazard, and each ground-truth hazard can be matched at most once. 
A match requires semantic consistency in the hazard category and object-level correspondence. Let $M$ denote the set of matched pairs returned by the judge. 
We compute
\begin{align}
    \mathrm{TP}_{hazard} &= |M|, \\
    \mathrm{FP}_{hazard} &= |\hat{\mathcal{H}}| - |M|, \\
    \mathrm{FN}_{hazard} &= |\mathcal{H}| - |M|.
\end{align}

\paragraph{Aggregation.}
For micro-averaged scores, we sum TP, FP, and FN over all evaluation samples before computing precision, recall, and F1. 
For macro-averaged scores, we first compute precision, recall, and F1 for each sample and then average the resulting scores across samples.

Category-level matching provides a lenient estimate because it ignores object-level grounding and may count a correct category as correct even when the reported object is wrong. 
Hazard-level matching is stricter because it requires object-level correspondence through one-to-one hazard matching.

\subsection{Benchmark Case}
\label{app:benchmark-case}

\begin{figure*}[t]
    \centering
    \includegraphics[width=\textwidth]{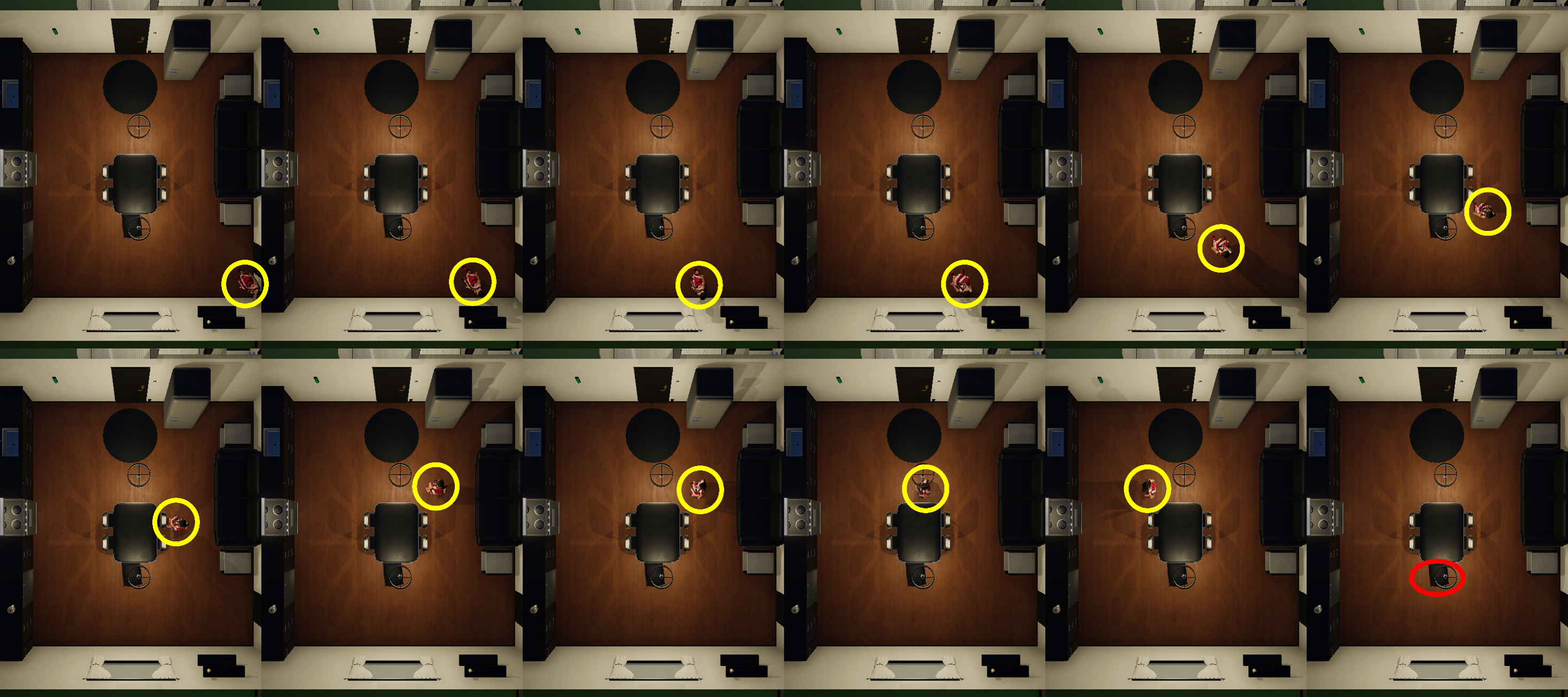}
    \caption{Top-down visualization of a Qwen3-VL-4B-Thinking agent interacting with the environment. Panels are ordered from left to right and top to bottom. Yellow circles mark the agent's current position in successive states, and the red circle in the final panel marks the ground-truth hazard for this task.}
    \label{fig:benchmark-case}
\end{figure*}

Figure~\ref{fig:benchmark-case} illustrates an example interaction between Qwen3-VL-4B-Thinking and the environment. To make the agent's movement through the room easy to follow, successive states are visualized from a top-down perspective and ordered from left to right and top to bottom. These top-down views are used only for visualization. During evaluation, the agent receives egocentric observations exclusively.

\section{Dataset Details}
\label{app:dataset-details}
\subsection{Annotation Details}

We first annotate potential hazard locations and object safety attributes in 12 room scenes from three VirtualHome environments. 
These annotations jointly specify where hazards may occur and which objects can instantiate them.
For hazard locations, annotators identify spatial regions that may become unsafe after an appropriate object is placed there, such as the top of a refrigerator, the interior of a sink, or the area near a stove.
Each location is associated with exactly one hazard type: fire, electric shock, falling object, trip, or child safety.
Two annotators each perform the initial annotation for six rooms across the three environments and then cross-check the other annotator's results case by case, removing low-risk locations and retaining only consensus annotations.
This process yields 136 annotated hazard locations.
For object attributes, two annotators independently determine whether common objects are flammable, electrical, likely to induce tripping or falling-object hazards, or unsafe for children, with multiple attributes allowed for each object.
Disagreements are adjudicated by a third annotator, resulting in 367 objects with safety-related attributes.

\begin{figure*}[t]
    \centering
    \begin{subfigure}[b]{0.9\textwidth}
        \centering
        \includegraphics[width=0.8\textwidth]{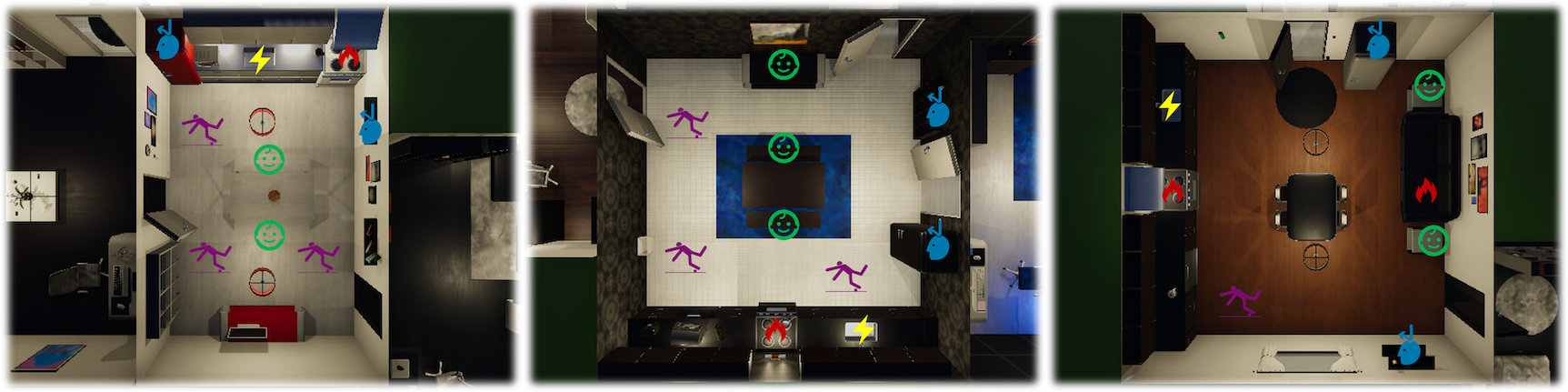}
    \end{subfigure}

    \begin{subfigure}[b]{0.9\textwidth}
        \centering
        \includegraphics[width=0.8\textwidth]{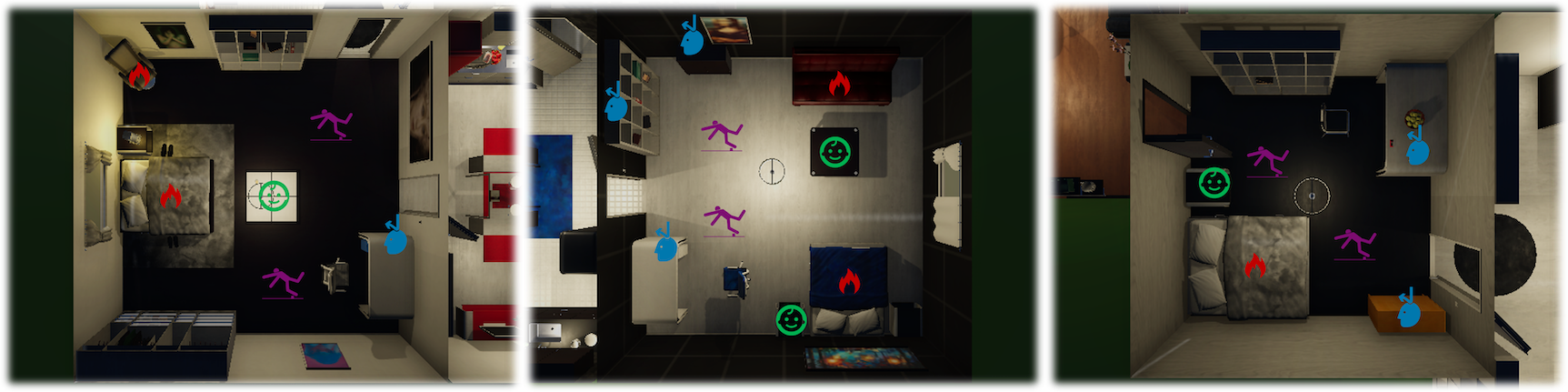}
    \end{subfigure}

    \begin{subfigure}[b]{0.9\textwidth}
        \centering
        \includegraphics[width=0.8\textwidth]{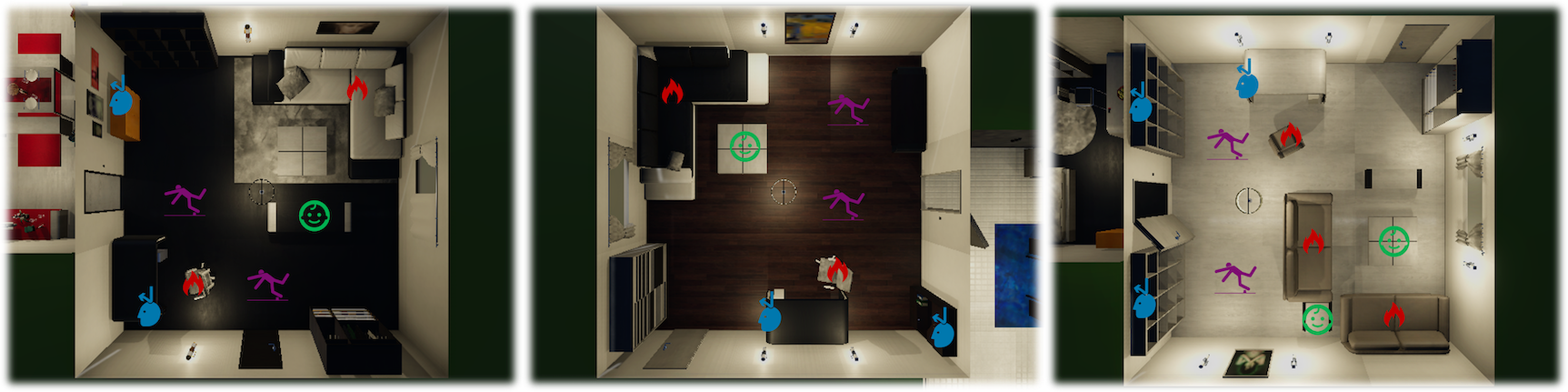}
    \end{subfigure}

    \begin{subfigure}[b]{0.9\textwidth}
        \centering
        \includegraphics[width=0.8\textwidth]{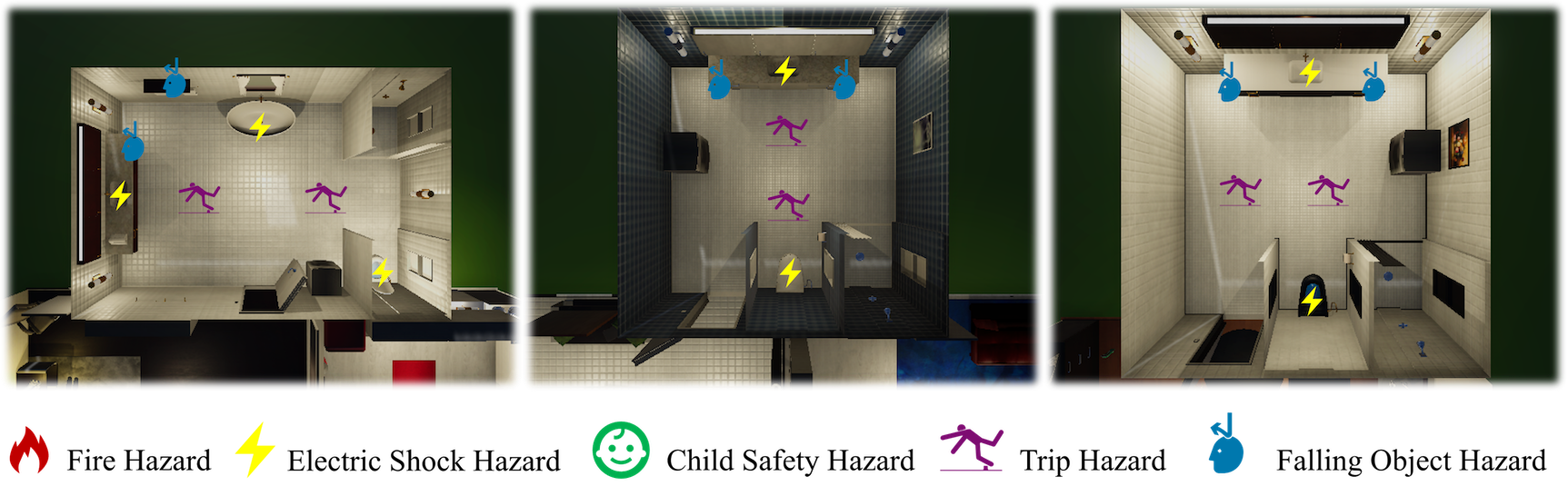}
    \end{subfigure}
    \caption{Examples of annotated hazard locations across the four room types, ordered from top to bottom as kitchen, bedroom, living room, and bathroom. Within each room type, the three scenes correspond from left to right to environments 0, 1, and 3.}
    \label{fig:hazard_position}
\end{figure*}

\paragraph{Hazard locations.}
We annotate candidate hazard locations for all four room types in each of the three environments.
Figure~\ref{fig:hazard_position} illustrates these annotations across kitchens, bedrooms, living rooms, and bathrooms.

\paragraph{Object attributes.}
Each placeable object is annotated with one or more safety-related attributes that determine the types of hazard locations with which it can be paired.
Table~\ref{tab:hazard-type-object-examples} lists example objects associated with each hazard type.

\begin{table}[ht]
  \centering
  \small
  \begin{tabular}{@{}p{0.22\columnwidth} p{0.68\columnwidth}@{}}
  \toprule
  Hazard type & Example objects \\
  \midrule
  Trip & Towel, box, bananas, dishwashing liquid, dish bowl \\
  Falling & Perfume, plate, mug, frying pan, orchid \\
  Fire & Towel, paper, perfume, box, pillow \\
  Children & Perfume, painkillers, cutlery knife, dishwashing liquid, condiment bottle \\
  Electric & Computer, radio, television, coffeemaker, microwave \\
  \bottomrule
  \end{tabular}
  \caption{Example objects associated with each hazard type according to their annotated safety attributes.}
  \label{tab:hazard-type-object-examples}
\end{table}

\subsection{VLM-Based Filtering Details}
For each rule-generated single-hazard candidate, we estimate a target center and spatial extent from the objects involved and render ten views around this region.
We define a local reference frame relative to the room interior and capture one frontal view, one top-down view, and eight oblique views sampled from four horizontal directions at upper and lower elevations.
Each camera is oriented toward the target center.
The capture distance is adapted to the spatial extent of the target region, and nominal camera placements are adjusted when possible to avoid intersections with scene geometry.

Walls, furniture, and local scene geometry may nevertheless cause individual views to be severely occluded or visually uninformative.
We therefore do not require all ten views to be valid.
The VLM jointly assesses the complete view set together with the intended hazard type and involved objects.
An isolated invalid view does not cause rejection when the remaining views provide sufficient evidence of the relevant objects, spatial relations, and scene context.
A candidate is retained only when the collective multi-view evidence supports both visual usability and semantic validity.
The complete filtering prompt is provided in Appendix~\ref{app:vlm-filter-prompt}.

\subsection{Single-Hazard Examples}
\begin{figure}[!htbp]
    \centering
    \includegraphics[width=1.0\linewidth]{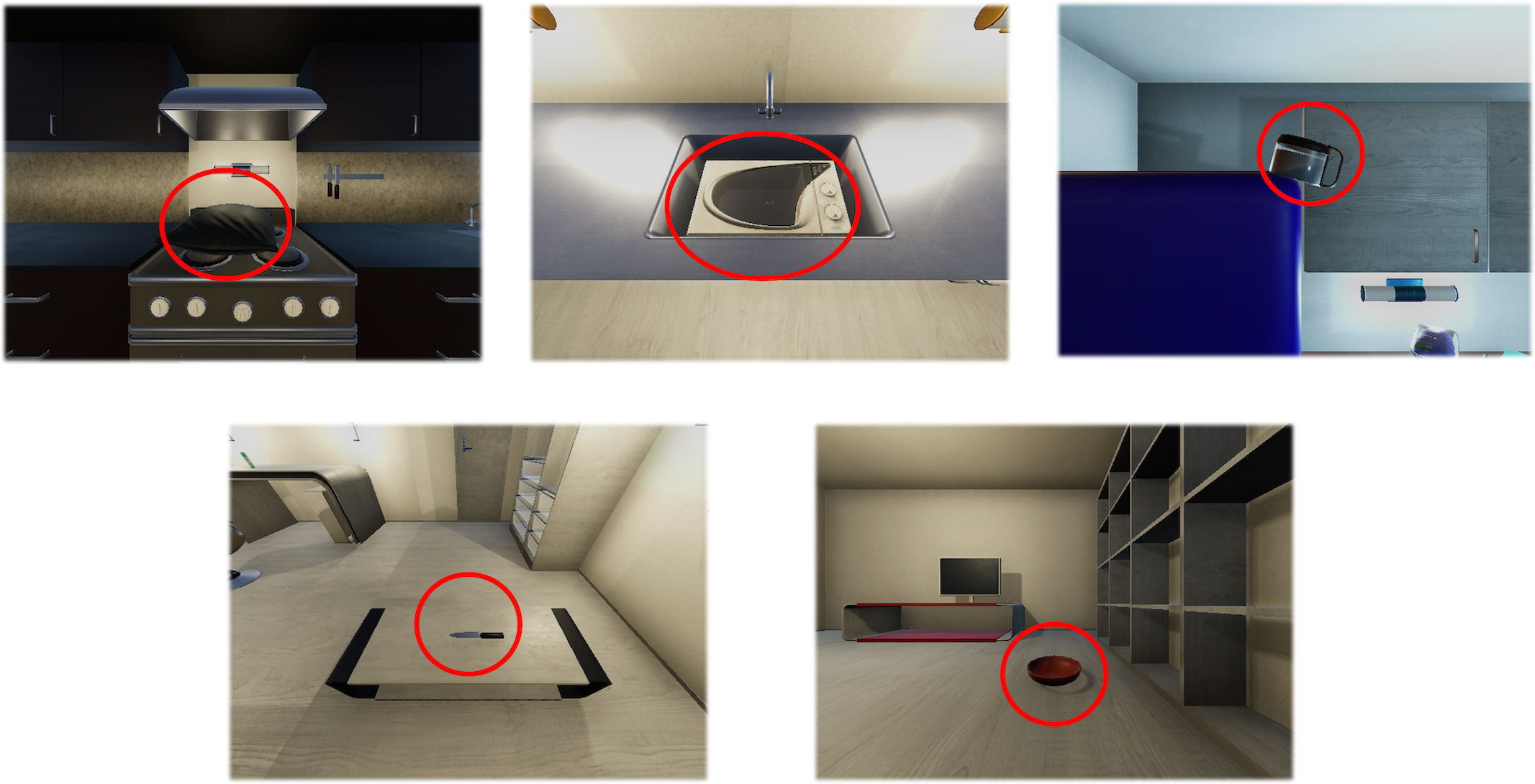}
    \caption{
    Single-hazard examples for the five hazard types.
    The first row shows a fire hazard (left), an electric-shock hazard (center), and a falling-object hazard (right).
    The second row shows a child-safety hazard (left) and a trip hazard (right).
    }
    \label{fig:type_hazard}
\end{figure}

Figure~\ref{fig:type_hazard} shows one retained single-hazard case for each of the five hazard types.
The examples depict a pillow placed on a stove (Fire), a microwave placed in a sink (Electric), a coffeepot positioned near the edge on top of a refrigerator (Falling), a knife left on a coffee table (Children), and a bowl placed in a walkway (Trip).

\begin{table*}[t]
  \centering
  \small
  \setlength{\tabcolsep}{5pt}
  \begin{tabular}{l c c c c c c}
  \toprule
  Split & \# Env. & \# Rooms & Loc. (Ann./Rep.) & SH cases & Tasks & Hazard inst. \\
  \midrule
  Train & 2 & 8 & 88/66 & 371 & 3,400 & 11,571 \\
  Test & 1 & 4 & 48/45 & 209 & 1,000 & 3,000 \\
  Total & 3 & 12 & 136/111 & 580 & 4,400 & 14,571 \\
  \bottomrule
  \end{tabular}
  \caption{Overview of the environment-disjoint data splits. Loc. (Ann./Rep.) reports all annotated locations and those represented by at least one retained single-hazard case. SH cases and Hazard inst. denote retained single-hazard cases and hazard instances in the composed tasks, respectively.}
  \label{tab:dataset-overview}
\end{table*}

\subsection{Dataset Statistics}
\label{app:dataset-statistics}

We use an environment-level split to prevent overlap in scene layouts and object instances.
The training set is constructed from environments 0 and 3, whereas the test set is constructed from the held-out environment 1.
The 580 VLM-retained single-hazard cases serve as reusable building blocks for composing data-level tasks.
Table~\ref{tab:dataset-overview} summarizes the resulting split sizes and location coverage.

We report hazard-type distributions separately for the retained single-hazard pool and the composed tasks.
Table~\ref{tab:single-hazard-type-distribution} counts unique retained cases.
Table~\ref{tab:composed-hazard-type-distribution} counts hazard instances after task composition.
Percentages are computed within each split.

The training set contains 371, 658, 657, 657, and 1,057 tasks with $N=1,2,3,4,5$ hazards, respectively.
The test set is balanced across hazard counts, with 200 tasks for each $N \in \{1,2,3,4,5\}$.

Table~\ref{tab:room-task-distribution} reports the task distribution across room types.
The test split is balanced with 250 tasks per room type.

\begin{table}[!htbp]
  \centering
  \footnotesize
  \setlength{\tabcolsep}{3.5pt}
  \begin{tabular}{@{}l r r r@{}}
  \toprule
  Hazard type & Train & Test & Total \\
  \midrule
  Trip & 142 (38.3\%) & 101 (48.3\%) & 243 (41.9\%) \\
  Falling & 119 (32.1\%) & 56 (26.8\%) & 175 (30.2\%) \\
  Fire & 73 (19.7\%) & 34 (16.3\%) & 107 (18.4\%) \\
  Children & 22 (5.9\%) & 13 (6.2\%) & 35 (6.0\%) \\
  Electric & 15 (4.0\%) & 5 (2.4\%) & 20 (3.4\%) \\
  Total & 371 (100\%) & 209 (100\%) & 580 (100\%) \\
  \bottomrule
  \end{tabular}
  \caption{Hazard-type distribution of retained single-hazard cases.}
  \label{tab:single-hazard-type-distribution}
\end{table}

\begin{table}[!htbp]
  \centering
  \footnotesize
  \setlength{\tabcolsep}{3.5pt}
  \begin{tabular}{@{}l r r r@{}}
  \toprule
  Hazard type & Train & Test & Total \\
  \midrule
  Trip & 3,929 (34.0\%) & 1,190 (39.7\%) & 5,119 (35.1\%) \\
  Falling & 3,316 (28.7\%) & 754 (25.1\%) & 4,070 (27.9\%) \\
  Fire & 2,361 (20.4\%) & 577 (19.2\%) & 2,938 (20.2\%) \\
  Children & 1,205 (10.4\%) & 305 (10.2\%) & 1,510 (10.4\%) \\
  Electric & 760 (6.6\%) & 174 (5.8\%) & 934 (6.4\%) \\
  Total & 11,571 (100\%) & 3,000 (100\%) & 14,571 (100\%) \\
  \bottomrule
  \end{tabular}
  \caption{Hazard-type distribution of instances in composed tasks.}
  \label{tab:composed-hazard-type-distribution}
\end{table}

\begin{table}[!htbp]
  \centering
  \small
  \begin{tabular}{l r r}
  \toprule
  Room type & Train & Test \\
  \midrule
  Bathroom & 577 & 250 \\
  Bedroom & 784 & 250 \\
  Kitchen & 1,255 & 250 \\
  Living room & 784 & 250 \\
  \midrule
  Total & 3,400 & 1,000 \\
  \bottomrule
  \end{tabular}
  \caption{Task distribution across room types.}
  \label{tab:room-task-distribution}
\end{table}

\section{CueBack Method Details}
\subsection{Action Trajectory Annotation and Synthesis}
\label{app:action-labeling}

\paragraph{Room-Level Graph Annotation.}
To construct executable teacher trajectories, we do not manually annotate a complete action sequence for every training task. Instead, we first build reusable navigation and observation annotations for each training room. Tasks instantiated in the same room share its spatial structure and potential hazard locations, allowing the same room-level annotations to support different numbers, types, and combinations of hazards.

For each training room, we instantiate the room, the character, and representative hazards at candidate locations in VirtualHome, and manually explore the scene. We record as graph nodes the positions from which one or more potential hazard locations can be inspected reliably. For each location covered by a node, we additionally annotate the inspection-action sequence required to bring the target into view. For example, inspecting a trip hazard on the floor typically requires looking down, whereas inspecting a falling hazard on top of a refrigerator may require turning and then looking up. Associating these target-specific inspection actions with reusable navigation positions allows the same node to serve multiple hazard targets. Some graph nodes are used only to connect navigable regions and are not associated with hazard inspection.

For two connectable graph nodes, annotators execute and record a navigation sequence that reliably reaches the target node in one direction. The corresponding reverse sequence is then derived automatically, so each manually annotated connection induces a pair of oppositely directed edges in the room graph.

The annotation of a room $r$ is represented as a directed graph $\mathcal{G}_r=(\mathcal{V}_r,\mathcal{E}_r)$, where $\mathcal{V}_r$ denotes the annotated graph nodes and $\mathcal{E}_r$ denotes executable directed navigation connections. Each node $v$ is associated with a coverage set $\mathcal{C}(v)$ containing the hazard locations that can be inspected from that node. For nodes used only for navigation connectivity, $\mathcal{C}(v)=\emptyset$.

\paragraph{Task-Conditioned Trajectory Synthesis.}
Given a training task, let $\mathcal{D}$ denote its set of target hazards. We match each target to graph nodes from which its location can be inspected. A target hazard may have multiple candidate inspection nodes, while one node may cover multiple target hazards. A complete inspection trajectory cannot be constructed if any target hazard has no valid inspection node.

A route starting from the initial character node is denoted by $\pi=(v_0,v_1,\ldots,v_T)$. A valid route follows navigation connections in the room graph and visits nodes that collectively cover all target hazards:
\begin{equation}
    \mathcal{D}
    \subseteq
    \bigcup_{t=0}^{T}\mathcal{C}(v_t).
    \label{eq:trajectory-coverage}
\end{equation}

We consider different inspection-node assignments and visitation orders, and connect the initial position and selected nodes using shortest paths in the directed graph. For an edge $e$, its cost $c(e)$ is computed from the annotated navigation sequence, where a forward movement of $n$ simulator steps contributes $n$ and each turning action contributes one. The total cost of a route is therefore $c(\pi)=\sum_{t=1}^{T}c(v_{t-1},v_t)$, which is used to rank candidate routes.

We then compile each graph route into an executable tool-interaction trajectory. Graph edges and target-specific inspection sequences are translated into navigation and view-adjustment tool calls, which are interleaved with egocentric observations and the corresponding hazard-reporting calls. A termination call is appended after all target hazards have been inspected.

The resulting tool-interaction trajectory contains temporally aligned egocentric observations, tool calls, and real environment feedback, and is used directly as multimodal supervision.

\begin{figure*}[t]
    \centering
    \includegraphics[width=\textwidth]{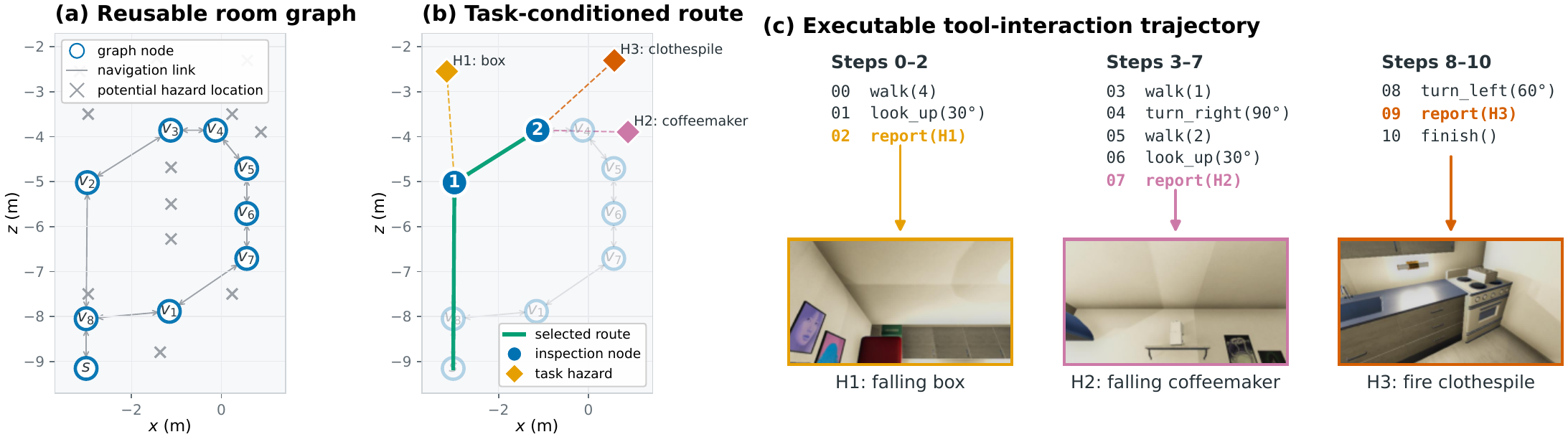}
    \caption{Illustration of task-conditioned trajectory synthesis. (a) Human annotations form a reusable room-level graph with executable navigation edges. Graph nodes associated with potential hazard locations serve as inspection nodes. For visual clarity, edges in (a) and (b) are drawn as straight lines. Each edge represents an executable navigation sequence that may include multiple movement and turning actions. (b) Given a task with three target hazards, inspection nodes covering the targets are selected and connected from the initial position. (c) The graph route is compiled into an executable tool-interaction trajectory, with each hazard-reporting call linked to its corresponding egocentric observation obtained during simulator replay.}
    \label{fig:trajectory-synthesis}
\end{figure*}

Figure~\ref{fig:trajectory-synthesis} illustrates the complete construction process using a training task with three target hazards, from the reusable room-level graph to the task-conditioned route and the replayed action trajectory.

\paragraph{Trajectory Replay and Validation.}
Because execution errors may accumulate when locally validated navigation connections are composed, the ranked candidate trajectories are replayed end to end and evaluated in terms of navigation reachability and visual validity. Navigation reachability requires the character to arrive within $1.0$ m of every target inspection node. Visual validity requires the target object to occupy at least 50 pixels in the instance-segmentation result at reporting time. We retain the first trajectory that satisfies both criteria, together with its complete action sequence, step-wise egocentric observations, and environment feedback.

\paragraph{Annotation and Trajectory Statistics.}
Across the eight rooms in the two training environments, we annotate 67 graph nodes, of which 60 have at least one valid hazard-inspection annotation. The room graphs contain 65 manually annotated navigation connections and 155 inspection annotations, covering 66 hazard locations used by the training tasks.

We first synthesize candidate action trajectories for 3,400 training tasks. After complete simulator replay, quality validation, and subsequent data processing, we obtain 3,158 executable training trajectories. For tasks containing $N=1,2,3,4,$ and $5$ target hazards, the resulting numbers of trajectories are 301, 598, 604, 603, and 1,052, respectively.

All retained trajectories satisfy the navigation-reachability and target-visibility requirements described above.

\subsection{Clue-Precedes-Confirmation Reasoning Annotation}

The action trajectories described above specify what the agent should do at each step, but they do not by themselves explain why the next action is appropriate. CueBack augments these executable trajectories with observation-grounded reasoning while keeping the original tool calls unchanged. The key idea is to make the reasoning follow the temporal structure of hazard discovery. Before a hazard clue is visible, navigation should be explained as broad exploration. Once the clue becomes visible, subsequent actions should be explained as deliberate approach or inspection. Only when the visual evidence is sufficient should the agent justify reporting the hazard.

Given a replayed trajectory, we first split it into pre-report segments. Each segment consists of the navigation and view-adjustment steps before a specific \texttt{report\_hazard} call, together with the hazard that will eventually be reported. For each segment, we ask GPT-5.5 to identify the earliest navigation step at which the reported hazard, or its clear risk area, becomes visually evident from the egocentric observations. This step is used as a boundary between two stages: steps before the boundary are labeled as \texttt{free\_explore}, while steps from the boundary until the report are labeled as \texttt{approach\_inspect}. The report step itself is labeled as \texttt{report}, and the final termination step is labeled as \texttt{finish}. If the hazard is not clearly visible in any pre-report navigation frame, the boundary is set to null and the segment remains in the free-exploration stage until reporting.

After obtaining these stage labels, we generate concise hidden reasoning for each assistant step. The reasoning-generation prompt receives the current observation image, the forced next tool call, the stage label, and the boundary decisions. It is required to justify the fixed action rather than choose a new action. For \texttt{free\_explore} steps, the reasoning describes broad room coverage or searching for useful views; for \texttt{approach\_inspect} steps, it describes moving or adjusting the view to inspect a visible clue more closely; for \texttt{report} steps, it explains why the visible evidence supports the specified hazard report; and for \texttt{finish} steps, it explains why inspection can terminate.

The generated reasoning is inserted into the assistant message before the original tool call, using the same thinking-style format as the target model. Thus, CueBack changes only the supervision text preceding each action, not the action trajectory itself. Compared with directly prompting a VLM to rationalize each action independently, this two-stage procedure encourages reasoning in which visual clues precede hazard confirmation, producing supervision that is better aligned with free-exploration safety inspection.

\section{Experiment Details}

\subsection{Implementation Environment}
All experiments are implemented on top of the VirtualHome simulator and the Unity rendering backend. We use the same simulator wrapper for data collection, training-data replay, and benchmark evaluation, so that all methods interact with the environment through an identical set of native tools for navigation, view control, and hazard reporting. Commercial VLMs are accessed through their official or OpenAI-compatible APIs, while open-source VLMs are served locally through an OpenAI-compatible inference server. For fine-tuning experiments, we use PyTorch and Hugging Face Transformers with distributed data parallel training. The simulator, model server, and evaluation runner are launched as separate processes, and each evaluation worker uses an independent simulator connection to avoid cross-episode state leakage.

\subsection{SFT Data Formatting}
\label{app:experiment-details-sft-data}
We convert each generated exploration trajectory into turn-level supervised examples. Each example contains the current first-person observation, the task instruction, and the interaction history before the target action. The prediction target is the next assistant response, represented as either an executable tool call or a reasoning-augmented response followed by the corresponding tool call. During training, all instruction, observation, and history tokens are treated as context, and the loss is applied only to the target assistant span. This format lets all SFT variants share the same action trajectories while differing only in whether and how observation-grounded reasoning is inserted before the action.

For \emph{SFT-Action}, the target assistant span contains only the executable action. For \emph{SFT-Both}, the target contains a rationale generated directly from the trajectory prefix and the current action. For CueBack, the target contains the clue-based reasoning produced by our clue-precedes-confirmation construction, followed by the same executable action. Thus, differences between the SFT baselines come from the supervision text rather than from different environment rollouts.

\subsection{SFT Training Details}
\label{app:experiment-details-sft-training}
All supervised fine-tuning experiments use Qwen3-VL-4B-Thinking as the base model. We apply LoRA adapters to both attention and feed-forward layers, including the query, key, value, output, gate, up, and down projection modules. The hyper-parameters are listed in the main paper. The visual encoder is kept frozen, and only the LoRA parameters in the language model are updated.

\subsection{DPO Implementation}
\label{app:experiment-details-dpo}
The DPO baseline uses the same base model, environment trajectories, and model-interface format as the SFT experiments, but replaces next-action supervision with offline preference optimization. For each task, we sample four complete rollouts from Qwen3-VL-4B-Thinking. The rollouts are ranked primarily by hazard-level F1, with ties broken in favor of shorter trajectories and token length. We then construct a preference pair by treating the best rollout as the chosen response and the worst rollout as the rejected response.

DPO is trained with the same LoRA-based fine-tuning stack used for SFT, and $\beta$ is set oto $0.1$. The model is optimized on the offline preference pairs without additional human annotation. As with SFT, the trained LoRA adapter is merged into the base checkpoint before benchmark evaluation.

\section{Use of Prompts}
\subsection{VLM-Based Filtering Prompt}
\label{app:vlm-filter-prompt}
For each single-hazard candidate, the placeholders in the following prompt are
instantiated with its hazard definition, hazard type, and involved objects. The
prompt is provided with all ten rendered views described in the main paper.

\begin{tcolorbox}[
title=Prompt for VLM-Based Hazard Filtering,
colback=gray!5,
colframe=gray!60,
boxrule=0.5pt,
arc=2mm,
breakable,
left=2mm,
right=2mm,
top=1mm,
bottom=1mm
]
\small
You are a strict annotation verifier. Treat the text Hint as a claim to be verified. Given multi-view images for one home-safety hazard, decide whether the images clearly support this Hint.

\medskip
\textbf{Background.}
\par\smallskip\noindent
The images come from a home inspection scenario. The dataset contains several types of potential hazards, such as fire hazards, electric hazards, falling-object hazards, trip hazards, and child-safety hazards.

\medskip
\textbf{Input.}
\par\smallskip\noindent
1. Multi-view images: all images show the same hazard from different viewpoints, covering the target objects, supporting surface or location, and surrounding context.
\par\smallskip\noindent
2. Hint: an annotation claim to be verified, including the hazard definition, hazard type, and the objects that form the hazard. The Hint may be wrong and must not be treated as fact.

\medskip
\textbf{Core verification task.}
\par\smallskip\noindent
Check whether the images prove that the Hint is correct. Keep a sample only when the images themselves clearly show the key objects named in the Hint and their spatial relation clearly matches the hazard type specified by the Hint.
\par\smallskip\noindent
If the Hint cannot be seen from the images, or if recognizing the objects, understanding the relation, or confirming the hazard requires relying on the Hint, delete the sample.
\par\smallskip\noindent
If the object identity, object combination, hazard type, or spatial relation in the images is inconsistent with the Hint, delete the sample.
\par\smallskip\noindent
Even if the images match the Hint, do not assign a high score if you independently judge that the home-safety risk is not significant, is weak, is forced, or is only theoretically possible.

\medskip
\textbf{Strict selection principle.}
\par\smallskip\noindent
The goal is not to decide whether the sample is barely usable. The goal is to select high-quality, strong-evidence samples that can be directly used for training or evaluation. Apply a strict standard.
\par\smallskip\noindent
The default tendency should be to filter out samples. Keep a sample only when the multi-view images provide clear, stable, and direct evidence.
\par\smallskip\noindent
Only output visual=usable or minor\_artifact and semantic=valid when all of the following are true:
\par\smallskip\noindent
1. The key hazard objects are clearly visible in at least two views.
\par\smallskip\noindent
2. The spatial relation between the key objects is clear and direct, without guessing.
\par\smallskip\noindent
3. The hazard semantics are obviously valid, not merely plausible.
\par\smallskip\noindent
4. There is no penetration, floating, severe occlusion, misplacement, abnormal scale, or unnatural contact that affects judgment.
\par\smallskip\noindent
5. The judgment is fully supported by visual evidence rather than completed or imagined from the Hint.
\par\smallskip\noindent
6. Looking at the images independently, the scene is also a significant, realistic, and safety-relevant home hazard, not a minor, forced, or low-risk case.
\par\smallskip\noindent
If key objects are small, blurry, occluded, only barely visible in one view, or the spatial relation is unclear, use unobservable or unusable.
\par\smallskip\noindent
If the hazard is only plausible, weakly supported, or requires relying on common sense or the Hint rather than the images, use uncertain or invalid.
\par\smallskip\noindent
When you hesitate between keep and reject/uncertain, choose the stricter label: unusable, unobservable, invalid, or uncertain.

\medskip
\textbf{View descriptions.}
\par\smallskip\noindent
- front: semantic front view of the hazard target.
\par\smallskip\noindent
- top: top-down view, useful for checking placement and obvious overlap.
\par\smallskip\noindent
- upper\_front\_right / upper\_front\_left / upper\_back\_right / upper\_back\_left: four high oblique views.
\par\smallskip\noindent
- lower\_front\_right / lower\_front\_left / lower\_back\_right / lower\_back\_left: four low oblique views.
\par\smallskip\noindent
Use all views together. Occlusion in a single view does not necessarily make the sample unusable.

\medskip
\textbf{Review goals.}
\par\smallskip\noindent
1. Visual usability: decide whether the hazard target objects are visible and whether there are obvious rendering or placement issues, such as penetration, severe overlap, floating, misplacement, abnormal scale, or severe occlusion.
\par\smallskip\noindent
2. Hint consistency and hazard semantics: decide whether the visible object identities, object combination, spatial relation, and scene context clearly support the hazard definition, hazard type, and hazard objects in the Hint.
\par\smallskip\noindent
3. Risk significance: even when the Hint is supported by the images, independently decide whether the risk is obvious, realistic, and practically meaningful for home safety.

\medskip
\textbf{Scores.}
\par\smallskip\noindent
\textbf{Visual quality score \texttt{visual\_score}:}
\par\smallskip\noindent
- 5: The key objects are clearly visible in at least two views, their placement is clear, and there is no penetration, floating, severe occlusion, misplacement, abnormal scale, or unnatural contact that affects judgment. Use 5 only for high-quality samples.
\par\smallskip\noindent
- 4: The key objects are mostly visible, with only minor visual artifacts or minor occlusion; judgment is still stable, but the sample is not perfect.
\par\smallskip\noindent
- 3: The key objects are visible but not clear enough, or there is moderate occlusion, slight misplacement, or slight scale issue; judgment requires caution.
\par\smallskip\noindent
- 2: The key objects are barely visible, or there are obvious visual issues such as notable penetration, overlap, floating, misplacement, or abnormal scale. The sample should not be kept.
\par\smallskip\noindent
- 1: The key objects are invisible or almost invisible, so visual review is impossible.

\medskip
\textbf{Hazard semantics score \texttt{semantic\_score}:}
\par\smallskip\noindent
- 5: The images independently, clearly, and directly support the Hint: all key objects named in the Hint are clearly identifiable, the object combination is correct, the spatial relation is explicit, and the hazard type strongly matches the hazard definition; additionally, looking at the images independently, the scene is a significant, realistic, safety-relevant home hazard. It does not rely on guessing or the Hint. Use 5 very conservatively.
\par\smallskip\noindent
- 4: The images mostly support the Hint, but one aspect is not strong enough, such as slightly uncertain object identity, a less direct relation, a need for minor context/common-sense support, or a risk that exists but is not significant or harmful enough. Do not keep.
\par\smallskip\noindent
- 3: The images only weakly support the Hint; the hazard may be valid but is unstable, the spatial relation is unclear, it mainly seems like this hazard because of the Hint, or the safety risk is minor/forced/low-risk. Do not keep.
\par\smallskip\noindent
- 2: The images are clearly not fully consistent with the Hint, for example one key object is missing, an object may be misidentified, or the object combination/spatial relation does not match; or the scene basically does not form a practically meaningful home-safety risk. Do not keep.
\par\smallskip\noindent
- 1: The images clearly do not support the Hint; key objects are missing/invisible, or the hazard type, object combination, or spatial relation conflicts with the Hint.

\medskip
\textbf{Rules.}
\par\smallskip\noindent
- The Hint is a claim to be verified and must not replace visual judgment. If the Hint conflicts with the images, trust the images and assign a low score.
\par\smallskip\noindent
- Do not reject only because one view is occluded. If other views clearly show the target and hazard, the sample can still be kept.
\par\smallskip\noindent
- Normal contact with a supporting surface or slight edge contact should not be treated as severe penetration.
\par\smallskip\noindent
- If the key objects named in the Hint cannot be found, or if the images do not clearly support the object relation specified by the Hint, do not guess; assign a low score.
\par\smallskip\noindent
- Do not assign a high score merely because the object categories match the Hint. If the risk is not obvious, serious, realistic, or is only a low-probability theoretical risk when judged from the images alone, \texttt{semantic\_score} must be at most 4.

\medskip
\textbf{Hint:}
\par\smallskip\noindent
\texttt{Hazard definition: \{HazardTypeDesc\}}
\par\smallskip\noindent
\texttt{Hazard type: \{HazardType\}}
\par\smallskip\noindent
\texttt{Hazard objects: \{HazardItems\}}

\medskip
\textbf{Output format:}
\par\smallskip\noindent
Output exactly one valid JSON object. Do not output markdown, explanations, or custom tags. All fields are required, and scores must be integers from 1 to 5.
\par\smallskip\noindent
\texttt{\{}
\par\smallskip\noindent
\quad\texttt{"visual\_score": 5,}
\par\smallskip\noindent
\quad\texttt{"semantic\_score": 5}
\par\smallskip\noindent
\texttt{\}}
\end{tcolorbox}

\subsection{Hazard-level VLM Judge Prompt}
For hazard-level matching, the VLM jointly considers all predicted and ground-truth hazards in a task. 
We use the following prompt to obtain a one-to-one assignment. Task-specific hazard descriptions have replaced the indicated placeholders.

\begin{tcolorbox}[
title=Prompt for Hazard Matching,
colback=gray!5,
colframe=gray!60,
boxrule=0.5pt,
arc=2mm,
breakable,
left=2mm,
right=2mm,
top=1mm,
bottom=1mm
]
\small
You are a senior evaluator for home safety inspection datasets.
\par\smallskip\noindent
You are strict about matching reports to the provided golden annotations and
do not invent hazards that are not listed.

\medskip
You will be given a list of predicted hazards and a list of golden hazards.
\par\smallskip\noindent
Each hazard includes a hazard type and the objects that cause the hazard.
\par\smallskip\noindent
Each golden hazard also includes a location, which briefly describes the
semantic location where the hazard occurs.

\medskip
Your task is to determine the best one-to-one matches between the predicted
hazards and the golden hazards.
\par\smallskip\noindent
A predicted hazard may also not match any golden hazard.

\medskip
\textbf{Predicted hazards:}
\par\smallskip\noindent
\texttt{\{predictions\}}

\medskip
\textbf{Golden hazards:}
\par\smallskip\noindent
\texttt{\{golden\_hazards\}}

\medskip

\textbf{Answer format:}
\par\smallskip\noindent
Return one entry for every predicted hazard.
\par\smallskip\noindent
Answer JSON only:
\par\smallskip\noindent
\texttt{\{}\newline
\quad\texttt{"predictions": [}\newline
\qquad\texttt{\{answer\_items\}}\newline
\quad\texttt{]}\newline
\texttt{\}}

\medskip
\emph{Gn} means one valid golden hazard ID from the provided golden hazard
list, such as G1, G2, \ldots, GN.
\par\smallskip\noindent
Use \texttt{null} when the predicted hazard does not match any golden hazard.
\par\smallskip\noindent
Do not return the literal string ``Gn.''

\medskip
\textbf{Judging rules:}
\par\smallskip\noindent
(1) Each predicted hazard can match at most one golden hazard, and each golden
hazard can match at most one predicted hazard.
\par\smallskip\noindent
(2) The hazard type must be the same or clearly semantically equivalent; exact
category-name equality is not required. For example, ``Child-Safety Hazard''
and ``children,'' ``Trip Hazard'' and ``trip,'' or ``Falling-Object Hazard'' and
``falling'' should be treated as equivalent.
\par\smallskip\noindent
(3) The reported objects and golden objects should be semantically the same,
closely related, or visually similar.
\par\smallskip\noindent
(4) Exact object-name equality is not required. For example, ``hair product''
and ``deodorant'' may both refer to visually similar small plastic bottles.
\par\smallskip\noindent
(5) The golden location may be used as supporting context.
\par\smallskip\noindent
(6) If the hazard type matches but the objects clearly refer to different
items, mark the prediction as unmatched.
\par\smallskip\noindent
(7) If multiple matches seem plausible, choose the best one-to-one matching.

\medskip
Do not include explanations, Markdown, or any text outside the JSON object.
\end{tcolorbox}

\subsection{Agent Prompt}
\label{app:agent-prompt}
The evaluated VLM agent receives a fixed system prompt at the start of each inspection episode. 
At every interaction step, the current egocentric image is then attached to a short user message with the step index. 
The tool interface is the one summarized in Table~\ref{tab:tool-schema}.
\begin{tcolorbox}[
title=Prompt for VLM-Based Inspection Agent,
colback=gray!5,
colframe=gray!60,
boxrule=0.5pt,
arc=2mm,
breakable,
left=2mm,
right=2mm,
top=1mm,
bottom=1mm
]
\small
\textbf{System prompt.}
\par\smallskip\noindent
You are a home safety inspection agent in a VirtualHome scene.
\par\smallskip\noindent
Use the provided tools to inspect the room step by step.
\par\smallskip\noindent
Call exactly one tool at a time.
\par\smallskip\noindent
Use \texttt{report\_hazard} when you identify visible hazards.
\par\smallskip\noindent
Use \texttt{walk}, \texttt{turn}, \texttt{look\_up}, or \texttt{look\_down} when
more inspection is needed.
\par\smallskip\noindent
Use \texttt{finish\_inspection} only when the current sample is complete.

\medskip
\textbf{Hazard category definitions:}
\par\smallskip\noindent
(1) Fire Hazard: Conditions that could start, intensify, or spread a fire.
\par\smallskip\noindent
(2) Electrical Hazard: Conditions involving electrical sources or powered items
that could cause shock, short circuit, overheating, or electrical damage.
\par\smallskip\noindent
(3) Falling-Object Hazard: Conditions where an item could drop, tip over, or
strike a person or nearby object.
\par\smallskip\noindent
(4) Trip Hazard: Conditions that could obstruct normal movement, reduce footing,
or cause a person to trip.
\par\smallskip\noindent
(5) Child-Safety Hazard: Conditions that could put a child at risk because a
child may access, misuse, ingest, touch, climb on, or become trapped by something
in the scene.
\par\smallskip\noindent
Report a category only when the visible scene provides plausible evidence of
that risk, not merely because an object type can be dangerous in general.
\par\smallskip\noindent
For this step, reason briefly at most 3 short sentences. Then immediately call exactly one tool. Do not spend the whole response on reasoning, the tool call is required.

\medskip
\textbf{Per-step user message.}
\par\smallskip\noindent
\texttt{Observation step \{step\}. Inspect the attached first-person image and call exactly one tool.}
\end{tcolorbox}

\subsection{Reasoning Generation Prompts}
\label{app:reasoning-generation-prompts}

The \emph{SFT-Both} baseline uses the first prompt below to directly generate concise reasoning for each forced action from the trajectory prefix and current observation. In contrast, CueBack uses a two-stage procedure. It first applies the boundary-localization prompt to identify the earliest step where the eventual hazard becomes visually clear enough to shift from free exploration to deliberate inspection, and then applies the stage-aware reasoning prompt to generate reasoning conditioned on the resulting stage labels.

\begin{tcolorbox}[
title=Prompt for Action-Both Reasoning Baseline,
colback=gray!5,
colframe=gray!60,
boxrule=0.5pt,
arc=2mm,
breakable,
left=2mm,
right=2mm,
top=1mm,
bottom=1mm
]
\small
\textbf{System message.}
\par\smallskip\noindent
You write concise hidden reasoning for a home-safety inspection agent.
\par\smallskip\noindent
For each provided observation image and forced next tool action, generate the assistant's reasoning that would naturally lead to that exact action.

\medskip
\textbf{Requirements:}
\par\smallskip\noindent
- Return JSON only: \texttt{\{"reasonings": ["..."]\}}.
\par\smallskip\noindent
- One reasoning per step, in the same order as the forced actions.
\par\smallskip\noindent
- Each reasoning must be short: 1--2 sentences, at most 45 words.
\par\smallskip\noindent
- Mention visible evidence when relevant, but do not invent precise details that are not visible.
\par\smallskip\noindent
- Do not include tool-call JSON, markdown, numbering, or the literal \texttt{</think>} marker.
\par\smallskip\noindent
- The action is fixed; justify it, do not choose a different action.

\medskip
\textbf{User message template.}
\par\smallskip\noindent
Task context: the agent is inspecting a VirtualHome room for hazards. It must call exactly one tool per step.
\par\smallskip\noindent
Below are the chronological observations and the forced next action for each step. Generate concise reasoning for every step.

\medskip
For each step, the following text is provided together with the corresponding observation image:
\par\smallskip\noindent
\texttt{Step \{step\}}
\par\smallskip\noindent
\texttt{User instruction: \{user\_instruction\}}
\par\smallskip\noindent
\texttt{Forced next action: \{tool\_call\}}
\end{tcolorbox}

\begin{tcolorbox}[
title=Prompt for CueBack Boundary Localization,
colback=gray!5,
colframe=gray!60,
boxrule=0.5pt,
arc=2mm,
breakable,
left=2mm,
right=2mm,
top=1mm,
bottom=1mm
]
\small
\textbf{System message.}
\par\smallskip\noindent
You label navigation stages for a home-safety inspection trajectory.
\par\smallskip\noindent
For each segment, the agent will eventually call \texttt{report\_hazard} for a specific hazard. Given the chronological observation images before that report, decide the earliest navigation step where the reported hazard or its clear risk area is basically visible enough that navigation should be explained as deliberate closer inspection rather than free exploration.
\par\smallskip\noindent
Return JSON only with this schema:
\par\smallskip\noindent
\texttt{\{"segments":[\{"report\_step":int,
"boundary\_step":int|null,
"rationale":"short"\}]\}}.

\medskip
\textbf{Rules:}
\par\smallskip\noindent
- \texttt{boundary\_step} must be one of that segment's navigation step numbers, or \texttt{null} if the hazard is not basically visible in any navigation image before the report.
\par\smallskip\noindent
- Before \texttt{boundary\_step}: free exploration. From \texttt{boundary\_step} onward until the report: deliberate approach/detail inspection.
\par\smallskip\noindent
- Do not choose a boundary just because the room type suggests risk; it must be supported by visible evidence in the images.

\medskip
\textbf{User message template.}
\par\smallskip\noindent
\texttt{Sample id: \{sample\_id\}. Label exactly one pre-report navigation segment.}
\par\smallskip\noindent
\texttt{Report step: \{report\_step\}}
\par\smallskip\noindent
\texttt{Final hazard to report: \{hazard\}}
\par\smallskip\noindent
\texttt{Navigation steps to label: \{nav\_steps\}}
\par\smallskip\noindent
\texttt{Return JSON only: \{"segments":[\{"report\_step":REPORT\_STEP,
"boundary\_step":STEP\_OR\_NULL,
"rationale":"short"\}]\}}

\medskip
For each candidate navigation step, the prompt includes:
\par\smallskip\noindent
\texttt{Navigation step \{step\}, forced action \{tool\_call\}. Observation image:}

\medskip
The report-step image is also attached as reference:
\par\smallskip\noindent
\texttt{Report step \{report\_step\} reference image immediately before report\_hazard:}
\end{tcolorbox}

\begin{tcolorbox}[
title=Prompt for CueBack Reasoning Generation,
colback=gray!5,
colframe=gray!60,
boxrule=0.5pt,
arc=2mm,
breakable,
left=2mm,
right=2mm,
top=1mm,
bottom=1mm
]
\small
\textbf{System message.}
\par\smallskip\noindent
You write concise hidden reasoning for a home-safety inspection agent.
\par\smallskip\noindent
For each observation image and forced next tool action, generate the assistant's reasoning that naturally leads to that exact action while respecting the provided stage label.
\par\smallskip\noindent
Return JSON only: \texttt{\{"reasonings":["..."]\}}.

\medskip
\textbf{Requirements:}
\par\smallskip\noindent
- One reasoning per step, same order as the forced actions.
\par\smallskip\noindent
- 1--2 short sentences, at most 45 words.
\par\smallskip\noindent
- Do not include tool-call JSON, markdown, numbering, or the literal \texttt{</think>} marker.
\par\smallskip\noindent
- The action is fixed; justify it, do not choose a different action.

\medskip
\textbf{Stage semantics:}
\par\smallskip\noindent
- \texttt{free\_explore}: the eventual hazard is not yet clear; explain navigation as broad room coverage / searching for useful views.
\par\smallskip\noindent
- \texttt{approach\_inspect}: a possible hazard or target area is visible; explain navigation as deliberately moving/turning/looking to inspect it more closely.
\par\smallskip\noindent
- \texttt{report}: evidence is sufficient; explain why reporting the specified hazard is appropriate.
\par\smallskip\noindent
- \texttt{finish}: inspection is complete; explain why finishing is appropriate.

\medskip
\textbf{User message template.}
\par\smallskip\noindent
\texttt{Sample id: \{sample\_id\}. Generate stage-aware reasoning for these steps only. Boundary decisions: \{boundaries\}}

\medskip
For each step, the following text is provided together with the corresponding observation image:
\par\smallskip\noindent
\texttt{Step \{step\}}
\par\smallskip\noindent
\texttt{Stage label: \{stage\_label\}}
\par\smallskip\noindent
\texttt{Forced next action: \{tool\_call\}. \{hazard\_hint\}}
\par\smallskip\noindent
\texttt{Observation image:}
\end{tcolorbox}

\section{Additional Experiments}
In the main text, we report hazard-level precision, recall, and F1 as the primary metrics. This evaluation performs one-to-one matching between predicted hazards and ground-truth hazards, and therefore requires the model to identify not only the correct hazard category but also the corresponding objects and hazard instance. It is the stricter setting and better reflects whether an embodied agent has found the intended safety issue. As a complementary metric, we also report category-level precision, recall, and F1, which only compare the predicted and ground-truth hazard categories and do not rely on an LLM judge for object-level matching. Table~\ref{tab:homesafebench-category-average} shows that category-level scores are consistently higher, as expected, while the overall conclusion remains unchanged, that our proposed CueBack achieves the best F1 among the evaluated training methods.

\begin{table}[t]
\centering
\setlength{\tabcolsep}{1.5mm}
\begin{tabular}{lcccccc}
\toprule
\multirow{2}{*}{\textbf{Model}} & \multicolumn{3}{c}{\textbf{Hazard-level}} & \multicolumn{3}{c}{\textbf{Category-level}} \\
\cmidrule(lr){2-4} \cmidrule(lr){5-7}
    & P & R & F1 & P & R & F1 \\

\midrule
\rowcolor{gray!20} \multicolumn{7}{c}{\it Commercial VLMs}\\
Seed-2.0-Pro & 41.4 & 29.9 & 34.7 & 66.4 & 48.2 & 55.8 \\
GPT-5.5 & 30.4 & 31.2 & 30.8 & 49.3 & 52.0 & 50.6 \\
GLM-5V-Turbo & 58.9 & 17.8 & 27.3 & 80.0 & 24.2 & 37.2 \\
Qwen-3.6-Plus & 45.7 & 19.3 & 27.1 & 68.3 & 28.8 & 40.5 \\
Kimi-K2.5 & 35.8 & 14.0 & 20.1 & 59.6 & 23.4 & 33.6 \\

\midrule
\rowcolor{gray!20} \multicolumn{7}{c}{\it Open-Source VLMs}\\
Qwen3-VL-8B & 40.2 & 14.2 & 21.0 & 66.2 & 23.3 & 34.5 \\
Qwen3-VL-4B & 47.2 & 11.6 & 18.7 & 71.8 & 17.9 & 28.6 \\
Qwen3-VL-A3B & 22.7 & 1.4 & 2.6 & 56.8 & 3.5 & 6.6 \\

\midrule
\rowcolor{gray!20} \multicolumn{7}{c}{\it Qwen3-VL-4B Trained with Our Data}\\
SFT-Action & 49.4 & 30.6 & 37.8 & 76.6 & 43.8 & 55.7 \\
SFT-Both & 55.7 & 30.4 & 39.3 & 84.7 & 46.2 & 59.7 \\
DPO & 40.6 & 22.7 & 29.1 & 56.1 & 31.4 & 40.3 \\
CueBack (ours) & 58.3 & 37.1 & 45.3 & 83.0 & 52.8 & 64.5 \\
\bottomrule
\end{tabular}
\caption{Average hazard-level and category-level results. Hazard-level scores are from Table~\ref{tab:homesafebench-main}, and category-level scores are computed without LLM-based hazard matching.}
\label{tab:homesafebench-category-average}
\end{table}

\section{Case Study}
\begin{figure*}[ht]
    \centering
    \includegraphics[width=\textwidth]{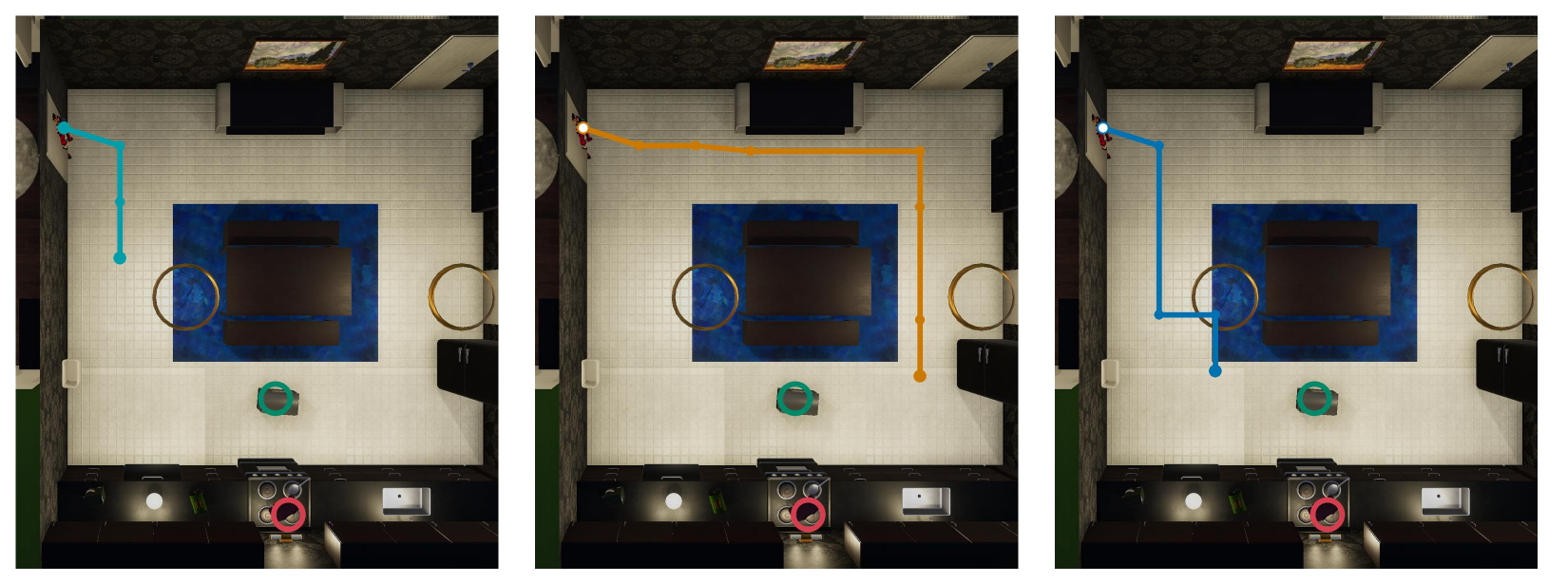}
    \caption{Top-down trajectory overview of the two-hazard kitchen inspection task. From left to right, the panels show the original VLM, the SFT-Both baseline, and \textsc{CueBack}. Colored paths mark the executed routes, and circles mark the two ground-truth hazards.}
    \label{fig:appendix-case-overview}
\end{figure*}

\begin{figure*}[ht]
    \centering
    \includegraphics[width=\textwidth]{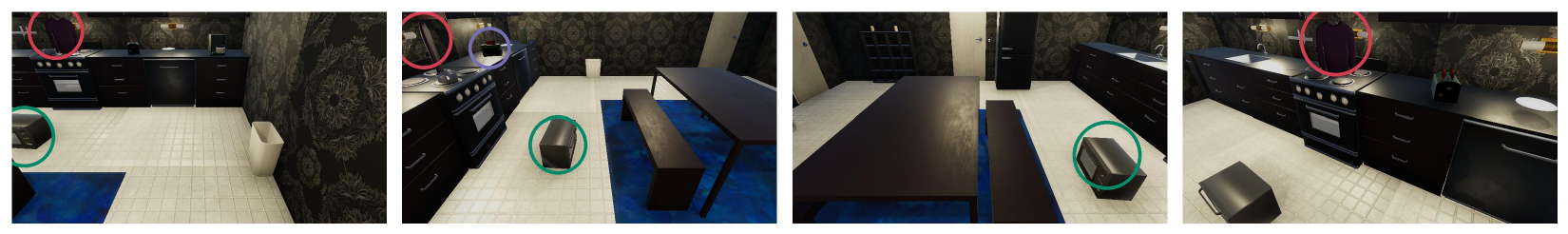}
    \caption{Representative egocentric observations from the three rollouts. The four views correspond to the original VLM before termination, the SFT-Both false-positive report, and the two correct reports made by \textsc{CueBack}. Green circles mark the microwave trip hazard, red circles mark the clothes-on-stove fire hazard, and purple circles mark the toaster false positive.}
    \label{fig:appendix-case-observations}
\end{figure*}

Figures~\ref{fig:appendix-case-overview} and~\ref{fig:appendix-case-observations} provide a qualitative comparison on a kitchen task with two target hazards. One hazard is a microwave placed in the walking area, which creates a trip risk. 
The other is a clothes shirt placed on the stove, which creates a fire risk. 
All compared agents are based on Qwen3-VL-4B. 
The original VLM follows a short sequence of looking down, walking, turning, and then terminating without a report. 
SFT-Both walks across the room, reports \texttt{trip / microwave}, continues moving, and later reports \texttt{trip / toaster}. \textsc{CueBack} walks toward the floor object, reports \texttt{trip / microwave}, then continues toward the stove area and reports \texttt{fire / clothes shirt}. 
This example separates two sources of error that are difficult to distinguish from aggregate metrics. 
The original model observes relevant regions but fails to report the hazards, while SFT-Both reports more actively but makes an object-grounding error.

\paragraph{Original VLM Case.}
The original VLM performs a short local inspection around the initial route, as shown in the left panel of Figure~\ref{fig:appendix-case-overview}. 
A representative egocentric view from this trajectory is shown in the first image of Figure~\ref{fig:appendix-case-observations}, where the stove area and floor region are already partially visible.
However, the model does not convert these visual cues into a hazard report. 
The episode ends with \texttt{finish\_inspection} and the stated reason that no visible hazards were detected. 
This behavior reflects a failure to sustain effective inspection. 
The model remains near the initial route and terminates before moving closer to the suspicious floor and stove regions, even though partial cues are already visible.

\paragraph{SFT-Both Case.}
The SFT-Both baseline explores farther, as shown in the middle panel of Figure~\ref{fig:appendix-case-overview}, and issues two hazard reports. 
It first reports \texttt{trip / microwave}, which matches one target hazard. Later, after additional turns and a short movement, it reports \texttt{trip / toaster}. 
The second image of Figure~\ref{fig:appendix-case-observations} shows this false-positive case. 
The report has the right general action pattern of stopping to report a potential floor-level risk, but it grounds the hazard to the wrong object and misses the clothes-on-stove fire hazard. 
In this case, SFT-Both becomes more willing to issue hazard reports, but its second report is not grounded to the correct object.

\paragraph{\textsc{CueBack} Case.}
\textsc{CueBack} follows a more direct inspection route, as shown in the right panel of Figure~\ref{fig:appendix-case-overview}, and reports hazards at observations where the relevant evidence is visible. 
The third and fourth images of Figure~\ref{fig:appendix-case-observations} show the two correct report views. 
The model first reports \texttt{trip / microwave}, then continues to the stove region and reports \texttt{fire / clothes shirt} before terminating. 
Compared with SFT-Both, \textsc{CueBack} makes the same number of reports, but both reports correspond to the target hazards. 
Its reports are made from views where the corresponding evidence is visible, and the reported objects match the target hazards. 
This behavior is consistent with the intended effect of CueBack supervision, which is to associate object-grounded reports with observations where the relevant visual evidence is available.

\end{document}